\documentclass[lettersize,journal]{IEEEtran}
\usepackage{amsmath,amsfonts}
\usepackage{algorithmic}
\usepackage{array}
\usepackage[caption=false,font=normalsize,labelfont=sf,textfont=sf]{subfig}
\usepackage{textcomp}
\usepackage{stfloats}
\usepackage{url}
\usepackage{cite}
\usepackage{verbatim}
\usepackage{graphicx}

\usepackage{algorithm}
\usepackage{listings}
\usepackage{etoolbox}
\makeatletter
\AfterEndEnvironment{algorithm}{\let\@algcomment\relax}
\AtEndEnvironment{algorithm}{\kern2pt\hrule\relax\vskip3pt\@algcomment}
\let\@algcomment\relax
\newcommand\algcomment[1]{\def\@algcomment{\footnotesize#1}}
\renewcommand\fs@ruled{\def\@fs@cfont{\bfseries}\let\@fs@capt\floatc@ruled
 \def\@fs@pre{\hrule height.8pt depth0pt \kern2pt}%
 \def\@fs@post{}%
 \def\@fs@mid{\kern2pt\hrule\kern2pt}%
 \let\@fs@iftopcapt\iftrue}
\makeatother
	
\usepackage{multirow}
\usepackage{wrapfig}
\usepackage{adjustbox}
\usepackage{amsmath, calc}
\usepackage{amssymb}
\usepackage{mathtools}
\usepackage{amsfonts}
\usepackage{mathrsfs}
\usepackage{nicefrac}       
\usepackage{microtype}      
\usepackage{color}
\usepackage{graphicx}
\usepackage{xcolor}
\usepackage{enumitem}
\usepackage{bbm}
\usepackage{dsfont}
\usepackage{enumitem}
\usepackage{array}
\usepackage{diagbox}
\usepackage{xspace}
\usepackage{booktabs}
\usepackage{comment}
\usepackage{hyperref}       
\usepackage{cleveref}
\usepackage{amsmath}
\usepackage{hhline}  

\usepackage{wrapfig}

\usepackage[utf8]{inputenc} 
\usepackage[T1]{fontenc}    

\usepackage{url}            
\usepackage{booktabs}       
\usepackage{amsfonts}       
\usepackage{nicefrac}       
\usepackage{microtype}      
\usepackage{xcolor}         
\makeatletter
\newcommand*\bigcdot{\mathpalette\bigcdot@{}}
\newcommand*\bigcdot@[2]{\mathbin{\vcenter{\hbox{\scalebox{#2}{$\m@th#1\bullet$}}}}}
\makeatother

\title{Prototypical Graph Contrastive Learning}

\definecolor{babyblueeyes}{rgb}{0.63, 0.79, 0.95}

\usepackage{lipsum}

\def\BibTeX{{\rm B\kern-.05em{\sc i\kern-.025em b}\kern-.08em
    T\kern-.1667em\lower.7ex\hbox{E}\kern-.125emX}}
\usepackage{balance}
\begin{document}
\title{Prototypical Graph Contrastive Learning}
\author{Shuai Lin, Chen Liu, Pan Zhou, Zi-Yuan Hu, Shuojia Wang, Ruihui Zhao, \\
Yefeng Zheng, Liang Lin, Eric Xing, and Xiaodan Liang
\IEEEcompsocitemizethanks{
        This work was supported in part by National Key R\&D Program of China under Grant No. 2020AAA0109700, National Natural Science Foundation of China (NSFC) under Grant No.61976233, Guangdong Province Basic and Applied Basic Research (Regional Joint Fund-Key) Grant No.2019B1515120039, Guangdong Outstanding Youth Fund (Grant No. 2021B1515020061), Shenzhen Fundamental Research Program (Project No. RCYX20200714114642083, No. JCYJ20190807154211365) and CAAI-Huawei MindSpore Open Fund. We thank MindSpore for the partial support of this work, which is a new deep learning computing framework (https://www.mindspore.cn).
        \emph{(Shuai Lin and Chen Liu have contributed to this work equally. Corresponding author: Xiaodan Liang.)}
        
		Shuai Lin, Chen Liu and Xiaodan Liang are with the School of Intelligent Systems Engineering, Sun Yat-sen University, Shenzhen 518107, China (e-mail: shuailin97@gmail.com; cathyliu41@gmail.com; xdliang328@gmail.com).
		
		Pan Zhou is with Sea AI Lab, Galaxis 138522, Singapore (e-mail: panzhou3@gmail.com).
		
		Zi-yuan Hu and Liang Lin are with the School of Computer Science and Engineering, Sun Yat-sen University, Guangzhou 510006, China.
		
		Shuojia Wang, Ruihui Zhao and Yefeng Zheng are with the Tencent Jarvis Lab, Tencent building, Shennan Avenue, Shenzhen 518000, China (e-mail: skylawang@tencent.com; zacharyzhao@tencent.com; yefengzheng@tencent.com).
		
		Eric Xing is with School of Computer Science, Mohamed bin Zayed University of Artificial Intelligence, Masdar City, Abu Dhabi, UAE (e-mail: eric.xing@mbzuai.ac.ae).
		}}


\markboth{IEEE TRANSACTIONS ON NEURAL NETWORKS AND LEARNING SYSTEMS}%
{How to Use the IEEEtran \LaTeX \ Templates}

\maketitle

\begin{abstract}
Graph-level representations are critical in various real-world applications, such as predicting the properties of molecules. But in practice, precise graph annotations are generally very expensive and time-consuming. To address this issue, graph contrastive learning constructs instance discrimination task which pulls together positive pairs (augmentation pairs of the same graph) and pushes away negative pairs (augmentation pairs of different graphs) for unsupervised representation learning. However, since for a query, its negatives are uniformly sampled from all graphs, existing methods suffer from the critical sampling bias issue, i.e., the negatives likely having the same semantic structure with the query, leading to performance degradation. To mitigate this sampling bias issue, in this paper, we propose a Prototypical Graph Contrastive Learning (PGCL) approach. Specifically, PGCL models the underlying semantic structure of the graph data via clustering semantically similar graphs into the same group, and simultaneously encourages the clustering consistency for different augmentations of the same graph. Then given a query, it performs negative sampling via drawing the graphs from those clusters that differ from the cluster of query, which ensures the semantic difference between query and its negative samples. Moreover, for a query, PGCL further reweights its negative samples based on the distance between their prototypes (cluster centroids) and the query prototype such that those negatives having moderate prototype distance enjoy relatively large weights. This reweighting strategy is proved to be more effective than uniform sampling. Experimental results on various graph benchmarks testify the advantages of our PGCL over state-of-the-art methods. Code is publicly available at https://github.com/ha-lins/PGCL.
\end{abstract}


\section{Introduction}
\IEEEPARstart{L}{earning} graph representations  is a fundamental problem in a variety of domains and tasks, such as molecular properties prediction in drug discovery~\cite{gilmer2017neural,chen2019graph}, protein function forecast in biological networks~\cite{alvarez2012new,jiang2017aptrank}, and community analysis in social networks~\cite{newman2004finding}. Recently, Graph Neural Networks (GNNs)~\cite{gilmer2017neural,kipf2016semi,xu2018powerful} have attracted a surge of interest and showed the effectiveness in learning graph representations. These methods are usually trained in a supervised fashion, which demands the task-specific labeled data. However, there are some aspects of shortcomings for the supervised training of GNN. Firstly, task-specific labels can be quite scarce for graph datasets (e.g.,  labeling biology and chemistry graph through human annotations are often resource-intensive). Secondly, due to the limited size of graph datasets, supervised GNNs are often confronted with the over-fitting and over-smoothing problems \cite{li2018deeper}, which limit their generalization capability to other tasks \cite{wu2021self}. Therefore, it is highly desirable to learn the transferable and generalizable graph representations in a self-supervised way on the large scale pre-training graph data. 
To this end, self-supervised approaches, such as generative
methods ~\cite{kipf2016variational,zhu2020selfsupervised} predictive methods ~\cite{hu2020pretraining} and contrastive methods ~\cite{li2019graph,sun2019infograph,you2020graph}, are coupled with GNNs to enable the graph representation learning leveraging unlabelled data. The learned representations from well-designed self-supervised pretext tasks are then transferred to down-stream tasks. 
Inspired by the advances of contrastive learning in those domains, graph contrastive learning have been proposed and then attracted huge attention for graph representation learning.  

\begin{figure}
    \begin{center}
    \includegraphics[width=0.48\textwidth]{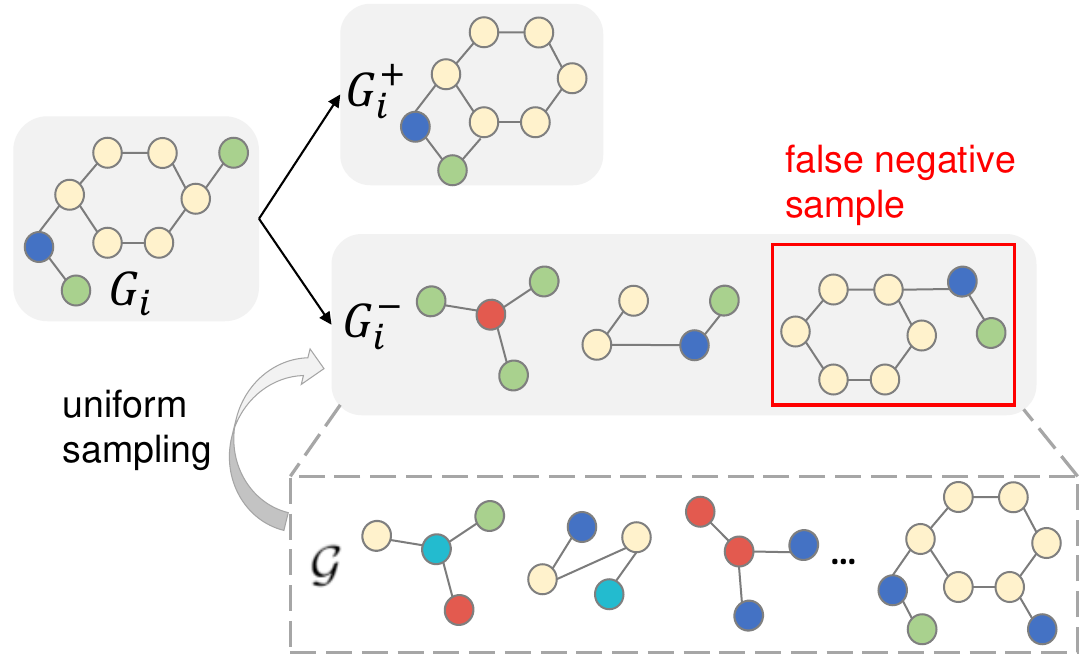}
    \caption{\textbf{``Sampling Bias''}: The strategy of sampling negative examples uniformly from the data distribution $\mathcal{G}$ could result in that the sampled negatives $G_i^{-}$ are semantically similar to the query $G_i$, e.g., they all contain the hexagonal structure that resembles a benzene ring.}
    \vspace{-15.5pt}
    \label{fig:sampling}
    \end{center}
\end{figure}

Graph contrastive learning is mainly based on maximizing the agreement of two views extracted from the same graph against those from different graphs. The former are regarded as positive pairs and later as negative ones. Specifically, three sequential components should be well-designed for graph contrastive learning, namely data augmentation, pretext task and contrastive objectives \cite{liu2021graph,wu2021self}. The first is the key for generating multiple appropriate views. Due to the inherent non-Euclidean
properties of graph data, it is difficult to directly apply data augmentations designed for images to graphs. Typical graph data augmentations includes shuffling node features  \cite{opolka2019spatial,ren2020heterogeneous}, perturbing structural connectivity through adding, masking and deleting nodes \cite{hu2020pretraining, zeng2020contrastive} and sub-graph sampling \cite{you2020graph} etc. The pretext task of graph contrastive learning contrasts two graph views at the same-scale or different scales. The scale of the view may be local, contextual, or global, corresponding to the node-level, subgraph-level, or graph-level information in the graph \cite{you2021graph, zhu2020deep,zhu2021adaptive,velivckovic2018deep ,jiao2020subgraph, sun2019infograph}. The main way to optimize the graph contrastive objective is maximizing the Mutual Information (MI) or the lower-bounds of MI (e.g., InfoNCE loss \cite{gutmann2010noise}) for two views. Typically, GraphCL~\cite{you2020graph} introduces four types of graph augmentations (namely node dropping, edge peturbation, attribute masking and subgraph sampling) and optimizes the InfoNCE loss on graph-level augmentations.

	However, all these graph contrastive methods suffer from the following limitations. Firstly, existing methods mainly focus on modeling the instance-level structure similarity but fail to discover the underlying global structure over the whole data distribution. But in practice,  there are  underlying global structures in the graph data in most cases. For example,  the graph MUTAG dataset~\cite{xu2018powerful}   is a dataset of mutagenic aromatic and heteroaromatic nitro compounds with seven discrete categories  which have underlying global structures but are not labeled to boost the representation learning.  
Secondly, as shown in Fig. ~\ref{fig:sampling}, for a query,  the common practice of sampling negatives uniformly from the whole data distribution could result in the fact that negatives  are actually semantically similar to the query.  However, these ``false'' negatives but really ``right'' positives  are undesirably pushed apart by the contrastive loss. This phenomenon, which we call   \emph{``sampling bias''}, can empirically lead to  significant performance degradation~\cite{chuang2020debiased}. Essentially, instance-wise contrastive learning learns an embedding space that only preserves the local similarity around each instance but largely ignores the global semantic structure of the whole graph data.

In this paper, we propose \emph{\textbf{p}rototypical \textbf{g}raph \textbf{c}ontrastive \textbf{l}earning} (PGCL), a new framework that clusters semantically similar graphs into the same group and simultaneously encourages the clustering consistency between different augmentations of the same graph. 
The global semantic structure of the entire dataset is depicted by PGCL in prototype vectors (i.e., trainable cluster centroids). 
Moreover, to address the sampling bias issue, we perform negative sampling via selecting the graphs from those clusters that differ from the query cluster. Specifically, we devise a reweighted contrastive objective, which reweights the negative samples based on the distance between their prototypes and the query prototype. In this way, those negative pairs having moderate prototype distance enjoy relatively large weights, which ensures the semantic difference between the query and its negative samples. 
In short,  the contributions of this paper can be summarized as follows:            
\begin{itemize}[leftmargin=*]
    \item We propose \emph{prototypical graph contrastive learning}, a novel framework that clusters semantically similar graphs into the same group and simultaneously encourages the clustering consistency between different augmentations of the same graph.
    \item We design a reweighted contrastive objective, which reweights the negative samples based on their prototype distance, to mitigate the \emph{sampling bias} issue. 
    \item Combining both technical contributions into a single model, PGCL outperforms instance-wise contrastive learning on multiple datasets in the task of unsupervised graph classification.
\end{itemize}

\section{Related Work}
\subsection{Graph Representation Learning}
\noindent Traditionally, graph kernels are widely used for learning node and graph representations. This common process includes meticulous designs like decomposing graphs into substructures and using kernel functions like the Weisfeiler-Leman graph kernel \cite{shervashidze2011weisfeiler} to measure graph similarity between them. However, they usually require non-trivial hand-crafted substructures and domain-specific kernel functions to measure the similarity while yielding inferior performance on downstream tasks, such as  node classification and graph classification. Moreover, they often suffer from poor scalability \cite{borgwardt2005shortest} and huge memory consumption \cite{kondor2016multiscale} due to some procedures like path extraction and recursive subgraph construction. Recently, there has been increasing interest in Graph Neural Network (GNN) approaches for graph representation learning and many GNN variants have been proposed \cite{ramakrishnan2014quantum, kipf2016semi, xu2018powerful}. 
A general GNN framework involves two key computations for each node at every layer: (1) AGGREGATE operation: aggregating
messages from neighborhood; (2) UPDATE operation: updating node representation from its representation in the previous layer and the aggregated messages. However, they mainly focus on supervised settings and differ from our unsupervised representation learning.     

\subsection{Contrastive Learning for GNNs}
\noindent There are some recent works that explored graph contrastive learning with GNNs in the aspects of data augmentations \cite{you2020graph,zhu2020graph}, pretext task designs \cite{qiu2020gcc,hu2020gpt} and contrastive objective \cite{hassani2020contrastive}. These methods can be mainly categorized into two types: global-local contrast and global-global contrast. The first category \cite{velivckovic2018deep,sun2019infograph,hassani2020contrastive,sun2021sugar, tong2021directed} follows the \textit{InfoMax principle} to maximize the the Mutual Information (MI) between the local feature and the context representation. Another line of graph contrastive learning approaches called global-global contrast \cite{you2020graph,zhang2020iterative,qiu2020gcc, fang2021molecular} directly studies the relationships between the global context representations of different samples as what metric learning does. Specifically, 
Deep Graph Infomax (DGI) \cite{velivckovic2018deep} adapted the idea from {Deep InfoMax} (DIM) to graphs for node representation learning via contrasting local node and global graph encodings with a summary vector. Further inspired by {Deep Graph Infomax}(DGI),  InfoGraph \cite{sun2019infograph} extends DIM to learn graph-level representations by maximizing the agreements between the representations of entire graphs and the representations of substructures of different scales (e.g., nodes, edges, triangles). MVGRL  \cite{hassani2020contrastive} trains the encoders through maximizing the mutual information from different structural views of graphs as well, while the same graph's adjacency and diffusion matrix were assumed as local and global views of a graph. Contrasting encodings between node representations of one view and graph representations of another view and vice versa yield better results compared with contrasting on local-local and global-global levels.  DiGCL \cite {tong2021directed} extends the constractive paradigm to directed graphs and aims at learning on abundant views while retaining the original structure information. Candidate views at topological- and feature-level are generated by a Laplacian perturbation operation on adjacency and node feature matrix. GNN-based encoder could therefore be adopted to the augmented veiws afterwards.
CGCN \cite{hui2020collaborative} explores whether unsupervised graph learning can boost the semi-supervised learning. In contrast, PGCL focuses on unsupervised graph contrastive learning.

More recently, MoCL \cite{sun2021mocl} and KCL \cite{fang2021molecular} both introduce the domain knowledge (e.g., manual rules and knowledge graph) into GCL and design knowledge-guided graph augmentation approaches to extract views. They boost the performance on the molecular graph since they have learned from the knowledge-guided augmented graphs. However, all these methods above are only able to model the discriminative relations between different graph instances while they fail to discover the underlying semantic structure of the data distribution. Meanwhile, randomly uniform negative sampling could leads to obtain the ``false'' negative pairs \cite{tschannen2019mutual, chuang2020debiased, lee2021augmentation}. This sampling bias phenomenon can empirically lead to significant performance degradation~\cite{chuang2020debiased}. Therefore, pondering how to sampling negative pairs with both structural and global semantic information carefully is the key process for graph contrastive learning. To mitigate the sampling bias issue, AFGRL \cite{lee2021augmentation} proposes an augmentation-free self-supervised method by merely generating positive pairs given a target node embedding for node-level tasks, while PGCL focus on the graph-level tasks. 
   

\subsection{Clustering-based Contrastive Learning}
\noindent Our work is also related to clustering-based representation learning methods \cite{li2018DynamicAG,caron2019deep,asano2019self,caron2019unsupervised,caron2020unsupervised,huang2019unsupervised,xu2021self,zhao2021graph,yuan2020self,li2018dynamic,li2020prototypical,zhou2021SLR,zhou2022mugs}. Among them, DeepCluster \cite{caron2019deep} and PCL \cite{li2020prototypical} show that K-means assignments can be used as pseudo-labels to learn visual representations. PCL\cite{li2020prototypical} introduces prototypes as latent variables to help find the maximum-likelihood estimation of the network parameters in an Expectation-Maximization framework, which encourages representations to be closer to their prototypes.
Other works \cite{asano2019self,caron2020unsupervised} show how to cast the pseudo-label assignment problem as an instance of the optimal transport problem. 
However, these methods are mainly developed for images instead of graph-structured data. Different data requires distinct solutions, e.g., data augmentations. 
In contrast, \cite{hui2020cgcn, xia2021selfsupervised, hu2021iter, liu2021multilayer, pan2021multi, jiang2021pre, hou2022neural} recently adapt the clustering idea to graph domain. 

Concretely, a self-supervised contrastive attributed graph clustering approach\cite{xia2021selfsupervised} is proposed to benefit from imprecise clustering labels for the node classification task. With stochastic graph augmentation schemes, augmented node attribute and topological graph structure are projected to low-dimensional vectors. Intra-cluster nodes were pulled together and inter-cluster nodes were pushed away in this process. Pre-GNN \cite{hu2021iter} designs a novel iterative feature clustering module that could be easily plugged into GCN. It's based on feature clustering and the pseudo labels predicted can both be updated in a EM-like style, which can further facilitate the node classification.  Liu et al. \cite{liu2021multilayer} proposes a multilayer graph contrastive clustering network, which clusters the nodes into different communities according to their relation types. Representations of the same node in different layers and different nodes were pulled closer and pushed away respectively by a contrastive objective. 
NCL \cite{lin2022improving} utilizes the cluster-based graph contrastive learning in the area of recommendation. 
For the multi-view attributed graph data, MCGC \cite{pan2021multi} learns the consensus graph by weighing different views and regularizes by graph contrastive loss. \cite{jiang2021pre} proposes graph pretraining approach for the heterogeneous graph, i.e., containing different types of nodes and edges. \cite{hou2022neural} utilizes the neural graph matching to pretrain graph neural network.
Concurrent to our work, GraphLoG \cite{xu2021self} also bring together a clustering objective with graph representation learning. Similar to PCL \cite{li2020prototypical}, GraphLoG applies K-means clustering to capture the graph semantic structure but utilizing K-means trivially could lead to imbalanced assignments of prototypes \cite{asano2019self}. Compared to GraphLoG, the proposed PGCL adds the constraint that the prototype assignments must be partitioned in equally-sized subsets and formulates it as an optimal transport problem. Moreover, PGCL aims to solve the sampling bias via sampling negatives from the clusters that differ from the query cluster and also reweighting negatives according to their prototype distances. 

\section{Preliminaries}
\subsection{Problem Definition}
A desirable representation should preserve the local similarity among different data instances. We give the more detailed discussions following \cite{xu2021self}:

\textbf{Local-instance Structure.}
We refer to the local pairwise similarity between various graph examples as the local-instance structure \cite{roweis2000nonlinear,xu2021self}. In the paradigm of contrastive learning, the embeddings of similar graph pairs are expected to be close in the latent space while the dissimilar pairs should be mapped far apart. 

The modeling of local-instance structure alone is usually insufficient to discover the global semantics underlying the entire data set. It is highly desirable to capture the global-semantic structure of the data, which is defined as follows:

\textbf{Global-semantic Structure.}
Graph data from the real world can usually be organized as semantic clusters \cite{furnas2017information,xu2021self}. The embeddings of nearby graphs in the latent space should embody the global structures, which reflect the semantic patterns of the original data.    

\textbf{Problem Setup.}\quad Given a set of unlabeled graphs $\mathcal{G} = \{G_i\}_{i=1}^{N}$, the problem of unsupervised graph representation learning aims at learning the low-dimensional vector $z_{i} \in \mathbb{R}^D$ of every graph $G_i \in \mathcal{G}$ which is favorable for downstream tasks, such as  graph classification.

\subsection{Graph Neural Networks}
In recent years, graph neural networks (GNNs) \cite{kipf2016semi,velivckovic2017graph,xu2018powerful} have emerged as a promising approach for learning representations of graph data. 
We represent a graph instance as $G = (\mathcal{V},\mathcal{E})$ with the node set $\mathcal{V}$ and the edge set $\mathcal{E}$. 
The dominant ways of graph representation learning are graph neural networks with neural message passing mechanisms \cite{hamilton2017inductive}:  at the k-th iteration/layer, node representation $ \mathbf{h}_{v}^{k}$ for every node $v \in \mathcal{V}$ is iteratively computed from the features of their neighbor nodes $\mathcal{N}(v)$ using a differentiable aggregation function. Specifically, at the iteration $k$ we get the embedding of node $v$ in the k-th layer as:
\begin{equation}
\footnotesize
\mathbf{h}_{v}^{k} = \textnormal{COMBINE}^{k}(\mathbf{h}_{v}^{k-1}, 
\textnormal{AGGREGATE}_{k}(\{\mathbf{h}_{u}^{k-1}, \forall u \in \mathcal{N}(v)\})) 
\label{eq:gnn}
\end{equation}
Then the graph-level representations can be attained by aggregating all node representations using a readout function, that is,
\begin{equation}
\label{eq:readout}
f_\theta(G_i) = \textnormal{READOUT}(\{\textnormal{CONCAT}(\{\mathbf{h}_{j}^{k}\}_{k=1}^{K})\}_{j=1}^{N})
\end{equation}
where $f_\theta(G_i)$ is the entire graph's embedding and READOUT represents averaging or a more sophisticated graph-level pooling function \cite{ying2018hierarchical,zhang2018end}.

\subsection{Graph Contrastive Learning}
To empower the GNN pre-training with unlabeled data, \textbf{g}raph \textbf{c}ontrastive \textbf{l}earning (GCL) has been explored a lot recently \cite{sun2019infograph,hassani2020contrastive,zhang2020iterative,you2020graph}.  
GCL performs pre-training through maximizing the agreement between two augmented views of the same graph via a contrastive loss in the latent space. GCL first augments the given graph to get augmented views $G_i$ and $G_i^\prime$, which are correlated (positive) pairs. Then $G_i$ and $G_i^\prime$ are fed respectively into a shared encoder $f_\theta$ (including GNNs and a following projection head) for extracting graph representations $z_i$, $z_i^\prime$ = $f_\theta(G_i),f_{\theta} (G_i^\prime)$
Then a contrastive loss function $\mathcal{L}(\cdot)$ is defined to enforce maximizing the consistency between positive pairs $z_i$, $z_i^\prime$ compared with negative pairs, such as InfoNCE loss~\cite{oord2018representation,hjelm2018learning,chen2020simple}:
\begin{equation}
\footnotesize
  \mathcal{L}_\mathbf{InfoNCE}=-\sum_{i=1}^{n}\log{\frac{\exp{\left(\boldsymbol{z}_i\cdot\boldsymbol{z}_i^\prime/\tau\right)}}{\exp{\left(\boldsymbol{z}_i\cdot\boldsymbol{z}_i^\prime/\tau\right)} + \sum_{j=1, j \neq i}^{2N}\exp{\left( \boldsymbol{z}_i\cdot\boldsymbol{z}_j/\tau\right)}}}
  \label{eq:infonce}
\end{equation}
where $z_i$ and $z_i^\prime$ are positive embeddings for graph $G_i$, and $z_j$ denotes the embedding of a different graph $G_j$ (i.e., negative embeddings), and $\tau$ is temperature hyper-parameter. Similar to \cite{xu2021self,arora2019theoretical}, in the graph-structured data, there is an underlying set of discrete latent classes $\mathcal{C}$ that represent semantic structures, which could result in that $G_i$ and $G_j$ are actually similar.       
 
\begin{figure*}[t]
	\setlength{\belowcaptionskip}{-0.1cm}
    \centering
    \includegraphics[width=\linewidth]{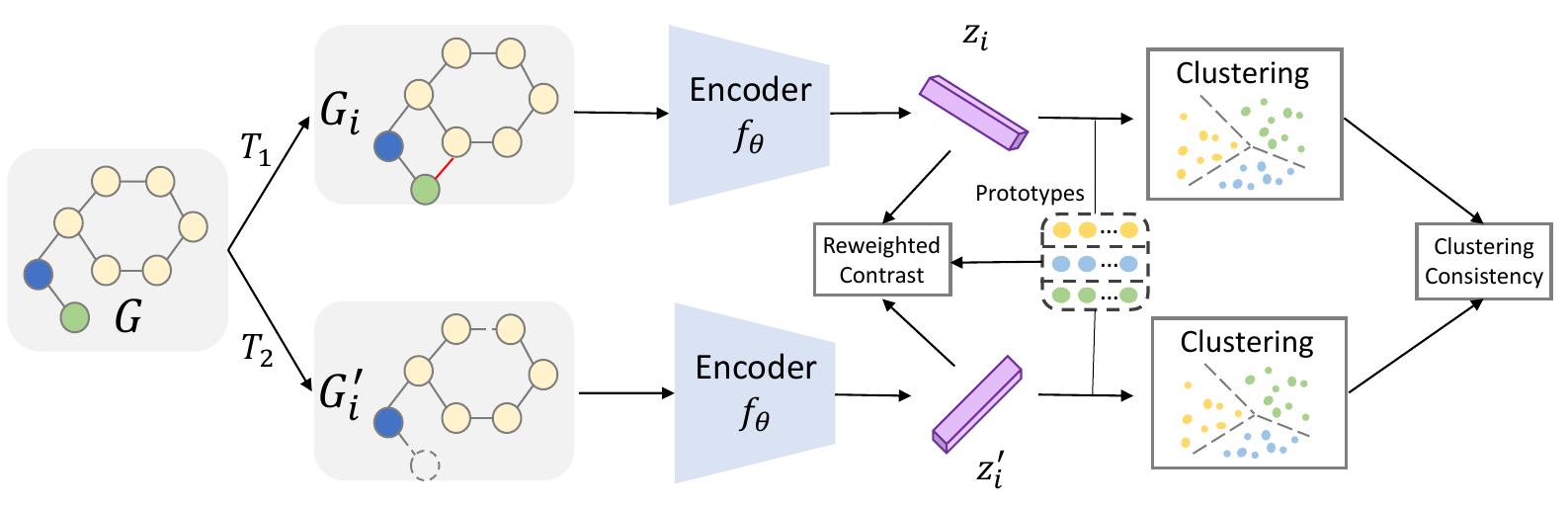}%
    \vspace{-0.1mm}
    \caption{\textbf{Overview of PGCL.} Two graph data augmentations $T_1$ and $T_2$ are applied to the input graph $G$. Then two graph views $G_i$ and $G^{\prime}_i$ are fed into the shared encoder $f_\theta$ (including GNNs and a projection head) to extract the graph representations $z_i$ and $z^{\prime}_i$. We perform the online clustering via assigning the representations of samples within a batch to prototype vectors (cluster centroids). The representations are learned via encouraging the clustering consistency between correlated views (Section~\ref{subsec:consistency}) and a reweighted contrastive objective (Section~\ref{subsec:reweighted}), where prototype vectors are also updated along with the encoder parameters by back-propragation.} 
    \label{fig:overview}
\end{figure*}

\section{Prototypical Graph Contrastive Learning}\label{method}

In this section, we introduce the Prototypical Graph Contrastive Learning (PGCL) approach. Our goal is to cluster semantically similar graphs into the same group and simultaneously encourage the clustering consistency between different augmentations of the same graph (i.e., correlated views). As shown in Fig.~\ref{fig:overview}, the representations of correlated views are encouraged to be clustered to have the same prototype (cluster centroid). Moreover, PGCL also designs a reweighted contrastive objective to sample the negatives from different clusters and reweight them according to their prototype distance.  
In the following, we introduce the PGCL in detail.

\subsection{Clustering Consistency for Correlated Views}
\label{subsec:consistency}
Formally, consider a graph neural network $z_i = f_\theta(G_i)$ mapping graph example $G_i$ to representation vectors $z_i\in\mathbb{R}^D$.
We can cluster all representations $z_i$ into $K$ clusters whose centroids are denoted by a set of $K$ trainable prototype vectors  $\{{c}_1,\dots,{c}_K\}$. Prototype vectors are trainable weight matrix of a feed forward network and initialized with He initialization~\cite{he2015delving}. For brevity, we denote by ${C} \in \mathbb{R}^{K \times D}$ the matrix whose columns are ${c}_1, \dots, {c}_K$. In practice, $C$ could be implemented by a single linear layer. In this way, given a graph $G_i$, we can perform clustering by computing the similarity between the representation $z_i=f_\theta(G_i)$ and the $K$ prototype as follows: 
\begin{equation}
p(y|z_i) = \mathrm{softmax} ({C}\cdot f_\theta(G_i)).
\end{equation}
Similarly, the prediction, i.e., $p(y|z^{\prime}_i)$, of assigning $G^{\prime}_i$ to prototypes can also be computed with its representation $z^{\prime}_i$. To encourage the clustering consistency between two correlated views $G_i$ and $G_i^\prime$, we predict the cluster assignments of $G_i^\prime$ with the representation $z_i$ (rather than $z^{\prime}_i$) from the correlated view and vice versa. Formally, we define the clustering consistency objective via minimizing the average cross-entropy loss:
\begin{equation}
\ell(p_i,q_{i^{\prime}})
=
-
\sum_{y=1}^K
q(y|z^{\prime}_i)
\log p(y|z_i)
\label{eq:centq}
\end{equation}
where $q(y|z^{\prime}_i)$ 
is the prototype assignment of the view $G^{\prime}_i$ and can serve as the target of the prediction $p(y|z_i)$ with $z_i$. The consistency objective acts as a regularizer to encourage the similarity of views from the same graph. We can obtain another similar objective if we swap the positions of $z_i$ and $z^{\prime}_i$ in Eqn.~\eqref{eq:centq} and the ultimate consistency regularizer can be derived by the sum of two objectives:  
\begin{equation}
\mathcal{L}_\mathbf{consistency}=\sum\nolimits_{i=1}^{n}\left[\ell(p_i,q_{i^\prime}) + \ell(p_{i^\prime},q_i)\right].
\label{eq:lcon}
\end{equation}

The consistency regularizer can be interpreted as a way of contrasting between multiple graph views by comparing their cluster assignments rather than their representations. In practice, optimizing the distribution $q$ faces the degeneracy problem since Eqn.~\eqref{eq:centq} can be trivially minimized by allocating all data samples to a single prototype. To avoid this, we add the constraint that the prototype assignments must be equally partitioned following \cite{caron2020unsupervised}. We calculate the objective in a minibatch manner for an efficient online optimization as:
\begin{equation}
 \footnotesize
  \min_{p,q} \mathcal{L}_\mathbf{consistency} \\
  \ \ \text{s.t.}\ \ 
  \forall y:q(y|z_i) \in \left[0,1\right] ~\text{and}~ \sum_{i=1}^N q(y|z_i) = \frac{N}{K}.
\label{eq:E_pq}
\end{equation}
The constraints mean that the prototype assignments to clusters $q(y|z_i)$ of each graph example $x_i$ are soft labels and that, overall, the $N$ graph examples within a minibatch are split uniformly among the $K$ prototypes. The objective in Eqn.~\eqref{eq:E_pq} is an instance of the \emph{optimal transport problem}, which can be addressed relatively efficiently. For more clarity, we denote two $K \times N$ matrices of joint probabilities as:
\begin{equation}
    P=\frac{1}{N}p(y|z_i); \text{  } Q=\frac{1}{N}q(y|z_i).
\label{eq:constraint}
\end{equation}
Then we can impose an equal partition by enforcing the matrix $Q$ to be a \emph{transportation polytope} following~\cite{asano2019self,caron2020unsupervised} in the minibatch manner:
\begin{equation}
  {T} = \left \{Q\in\mathbb{R}_+^{K\times N} ~|~Q \mathbbm{1}_N = \frac{1}{K} \mathbbm{1}_K, Q^\top \mathbbm{1}_K = \frac{1}{N} \mathbbm{1}_N \right \},
\end{equation}
where $\mathbbm{1}_N$ and $\mathbbm{1}_K$ denotes the vector of all ones with dimension of $N$ and $K$, respectively.
Then the loss function in Eqn. (\ref{eq:E_pq}) can be rewritten as:
\begin{equation}
  \min_{p,q} \mathcal{L}_\mathbf{consistency} 
  = 
  \min_{Q \in \mathbf{T}} \langle Q, - \log P \rangle - \log N,
\label{transport}
\end{equation}
where $\langle \cdot \rangle$ is the Frobenius dot-product between two matrices and $\log$ is applied element-wise. Optimizing Eqn.~\eqref{transport} always leads to an integral solution despite having relaxed $Q$ to the continuous polytope ${T}$ instead of the discrete one. We solve the transport problem
via utilizing the Sinkhorn-Knopp algorithm~\cite{cuturi2013sinkhorn} and the solution of
Eqn.~\eqref{transport} takes the form as:
\newcommand{\diag}{\operatorname{Diag}}
\begin{equation}
  Q = \diag(\alpha) P^\eta \diag(\beta)
\label{eq:final_q}
\end{equation}
where $\alpha$ and $\beta$ are two renormalization vectors and the exponentiation is element-wise. Here $\eta$ is chosen to trade off convergence speed with closeness to the original transport problem and it is a fixed value in our case. The renormalization vectors can be calculated using matrix multiplications with the Sinkhorn-Knopp algorithm~\cite{cuturi2013sinkhorn}. Note that the first term of Eqn. (10) is $\langle Q, - \log P\rangle$, while that is $\langle Q, P\rangle$ in Eqn. (2) of ~\cite{cuturi2013sinkhorn}. Thus the original exponential term is replaced with $P^\eta$.

\subsection{Reweighted Contrastive Objective}
\label{subsec:reweighted}
In this section, we introduce how to mitigate the sampling bias issue via sampling graphs from distinct  clusters to the query and reweighting the negative samples. In the image domain, some previous works \cite{chuang2020debiased,robinson2020contrastive} propose to approximate the underlying “true” distribution of negative examples by adopting a PU-learning viewpoint \cite{elkan2008learning}. However, such approximation is sensitive to the hyperparameter choice and cannot avoid sampling the semantically similar pairs essentially. Given a query (and its cluster), we can achieve this simply by drawing ``true'' negative samples from different clusters. Since different clusters represent distinct underlying semantics, such sampling strategy can ensure the semantic differences between the query and its negatives, and Eqn.~\eqref{eq:infonce} can be extended to:
\begin{equation}
\footnotesize
  \mathcal{L}=-\sum_{i=1}^{n}\log{\frac{\exp{\left(\boldsymbol{z}_i\cdot\boldsymbol{z}_{i}^\prime/\tau\right)}}{\exp{\left(\boldsymbol{z}_i\cdot\boldsymbol{z}_{i}^\prime/\tau\right)} + \sum_{j=1, j \neq i}^{2N} \mathbbm{1}_{{c}_{i} \neq {c}_{j}}\cdot\exp{\left( \boldsymbol{z}_i\cdot\boldsymbol{z}_j^\prime/\tau\right)}}}
  \label{eq:sampling_infonce}
\end{equation}
where ${c}_i$ and ${c}_j$ are the prototype vectors of graphs $G_i$ and $G_j$ respectively, and $\mathbbm{1}_{{c}_{i} \neq {c}_{j}}$ is the indicator that represents whether two samples are from different clusters. In this way, selected negative samples can enjoy desirable semantic difference from the query and those similar ones are ``masked'' out in the objective.

\begin{figure}
  \vspace{-15pt}
    \begin{center}
    \includegraphics[width=0.45\textwidth]{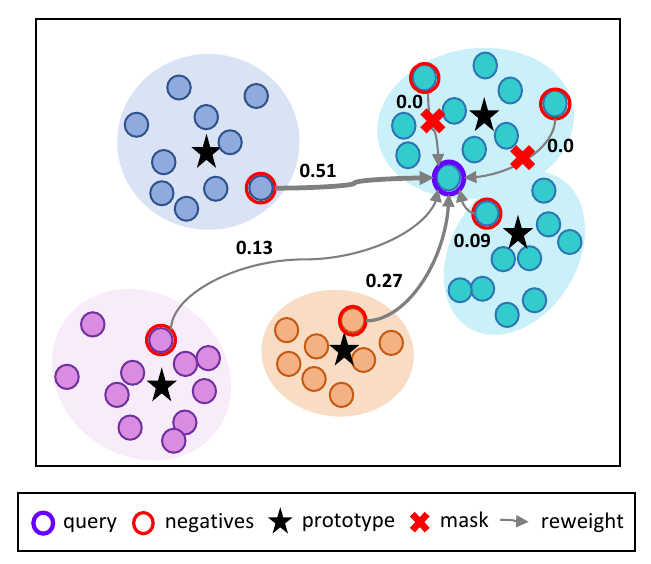}
    \caption{\textbf{Illustration of the negative sample reweighting.} The line width of the arrow denotes the weight value.}
      \vspace{-15.5pt}
    \label{fig:reweighted}
    \end{center}
\end{figure}

Beyond selecting negative samples based on the distinction of their clusters, we would like to avoid selecting too easy samples that are far from the query in the latent space. Furthermore, intuitively, the desirable negative samples should have a moderate distance to the query. Empirically, we found that controlling their prototype distances performs better than using their direct sample distance. As illustrated in Fig. \ref{fig:reweighted}, on the one hand, if the prototypes of negatives are too close to the query's prototype, negatives could still share the similar semantic structure with the query (e.g., the nearby cyan cluster). On the other hand, if the prototypes of negatives (such as the purple cluster) are far from the prototype of a query, it means negatives and query are far from each other and can be well distinguished, which actually does not help the representation learning. 

{
\small
\begin{align}
  & \mathcal{L}_\mathbf{Reweighted}=-\sum_{i=1}^{n} \\
  & \log{\frac{\exp{\left(\boldsymbol{z}_i\cdot\boldsymbol{z}_{i}^\prime/\tau\right)}}{\exp{\left(\boldsymbol{z}_i\cdot\boldsymbol{z}_{i}^\prime/\tau\right)} + M_i \sum_{j=1, j \neq i}^{2N} \mathbbm{1}_{{c}_{i} \neq {c}_{j}}\cdot\boldsymbol{w}_{ij}\cdot\exp{\left( \boldsymbol{z}_i\cdot\boldsymbol{z}_j^\prime/\tau\right)}}} \nonumber
  \label{eqn:one_self_loss}
\end{align}
}
where $w_{ij}$ is the weight of negative pairs ($G_i$, $G_j$) and $M_i = \frac{2N}{\sum_{j=1}^{2N}\boldsymbol{w}_{ij}}$ is the normalization factor. We utilize the cosine distance to measure the distance between two prototypes ${c}_{i}$ and ${c}_{j}$ as: 
    $\mathcal{D}({{c}_{i}, {c}_{j}}) = 1 - \frac{{c}_{i} \cdot {c}_{j}}{\left\|{c}_{i}\right\|_{2} \left\|{c}_{j}\right\|_2}$
. Then we define the weight based on the above prototype distance with the format of the Gaussian function as: 
\begin{equation}
\boldsymbol{w}_{ij}=  \exp\left\{-\frac{{	\left[
\mathcal{D}({{c}_{i}, {c}_{j}}) -\mu_{i} \right]^2}
}{2\sigma_{i}^{2}}\right\} 
  \label{eqn:total_self_loss}
\end{equation}
where $\mu_{i}$ and $\sigma_{i}$ are the mean and standard deviation of $\mathcal{D}({{c}_{i}, {c}_{j}})$ for query $G_i$, respectively.  

 \begin{algorithm}[t]
	\caption{Pseudocode of Prototypical Graph Contrastive Learning (PGCL) in Pytorch-like style.}
	\label{alg:code}
	\definecolor{codeblue}{rgb}{0.25,0.5,0.5}
	\lstset{
		backgroundcolor=\color{white},
		basicstyle=\fontsize{7.2pt}{7.2pt}\ttfamily\selectfont,
		columns=fullflexible,
		breaklines=true,
		captionpos=b,
		commentstyle=\fontsize{7.2pt}{7.2pt}\color{codeblue},
		keywordstyle=\fontsize{7.2pt}{7.2pt},
	}
	\begin{lstlisting}[language=python]
# C: prototypes (DxK)
# model: GIN + projection head
# temp: temperature
	
for x in loader:  # load a batch x with N samples
	G1 = T1(x)   # T1 is a random augmentation
	G2 = T2(x)  # T2 is a another random augmentation
	z = model(cat(G1, G2)) # embeddings: 2NxD
	
	scores = mm(z, C)  # prototype scores: 2NxK
        scores1 = scores[:N]
        scores2 = scores[N:]

        # cluster assignments
	with torch.no_grad():
	    q1 = sinkhorn(scores)
	    q2 = sinkhorn(scores2)
	    
	# convert scores to probabilities
	p1 = Softmax(scores1 / temp)
	p2 = Softmax(scores2 / temp)
	
	# clustering consistency loss
	Loss_Consistency = - 0.5 * mean(q2 * log(p1) + q1 * log(p2))
	
	# reweighted contrastive loss
	Compute Loss_Reweighted according to Eq.(13).
	
	# final loss
	loss = Loss_Reweighted + n * Loss_Consistency
	
	# SGD update: network and prototypes
	loss.backward()
	update(model.params)
	update(C)
	
	# normalize prototypes
	with torch.no_grad():
	    C = normalize(C, dim=0, p=2)
	\end{lstlisting}
\end{algorithm}

As shown in Fig. \ref{fig:reweighted}, the reweighting strategy encourages that lager weights are assigned to meaningful negative samples (such as from the blue and orange clusters) with a moderate prototype distance to the query and smaller weights to too easy negative samples (e.g., from the purple cluster) and ``false'' negative samples (from the nearby cyan cluster). The strategy is similar to those in \cite{zhao2014accelerating,lin2017focal} but they apply it on training samples under supervised learning while we adopt it for selecting negative samples of self-supervised learning. The final training objective couples $\mathcal{L}_\mathbf{Reweighted}$ and $\mathcal{L}_\mathbf{Consistency}$ as:
\begin{equation}
    \mathcal{L} = \mathcal{L}_\mathbf{Reweighted} + \lambda \mathcal{L}_\mathbf{Consistency}
  \label{eq:final_loss}
\end{equation}
where the constant $\lambda$ balances the reweighted contrastive loss $\mathcal{L}_\mathbf{Reweighted}$ and the consistency regularizer $\mathcal{L}_\mathbf{Consistency}$. This loss function is jointly minimized with respect to the prototypes
$\mathbf{C}$ and the parameters $\theta$ of the graph encoder used to produce the representation $z_i$.

\section{Experiments}
This section is devoted to the empirical evaluation of the PGCL approach. Our initial focus is on unsupervised learning. We further apply PGCL to the transfer learning setting to test the out-of-distribution performance. Finally, we perform the extensive experiments for analysis, including ablation studies, sensitivity analysis and visualization on the unsupervised learning datasets.

\subsection{Unsupervised Learning}
\paragraph{Task and datasets} We conduct experiments by comparing with the state-of-the-art competitors on the unsupervised graph classification task \cite{sun2019infograph,you2020graph}, where we only have access to all unlabeled samples in the dataset. We pre-train using the whole dataset to learn graph embeddings and feed them into a downstream SVM classifier with 10-fold cross-validation.
For this task, we conduct experiments on seven well-known benchmark datasets \cite{morris2020tudataset} including four bioinformatics datasets (MUTAG, PTC, PROTEINS, NCI1) and three social network datasets (COLLAB, RDT-B and RDT-M5K) with statistics summarized in Table \ref{tab:unsup_results}. 

\begin{table*}
\renewcommand\arraystretch{1.05}
\tabcolsep 0.1in 
\small
\centering
\begin{tabular}{@{}clccccccc@{}}\cmidrule[\heavyrulewidth]{2-9}
\multirow{3}{*}{\rotatebox{90}{\hspace*{0pt}Datasets}}
& Datasets & {\textsc{MUTAG}} & {\textsc{PTC}} & {\textsc{PROTEINS}}  & {\textsc{NCI1}} & {\textsc{COLLAB}} & {\textsc{RDT-B}} & {\textsc{RDT-M5K}}   \\
& \text{\# graphs }  & 188  & 344 & 1113 & 4110 & 5000  &  2000 & 2000 \\
& \text{Avg \# nodes }  &  17.9 & 14.3 & 39.1 & 29.9 &  74.5 &  429.6  & 429.6     
\\ \cmidrule{2-9}

\multirow{4}{*}{\rotatebox{90}{\hspace*{-8pt} Supervised }} 
& \textsc{GraphSAGE} \cite{hamilton2017inductive} &  85.1$\pm$7.6 &  63.9$\pm$7.7 &  75.9$\pm$3.2 & 77.7$\pm$1.5 & - & - & - \\
& \textsc{GCN} \cite{kipf2016semi} &  85.6$\pm$5.8 & 64.2$\pm$4.3 &  76.0$\pm$3.2 &  80.2$\pm$2.0 & 79.0$\pm$1.8 & 50.0$\pm$0.0 & 20.0 ± 0.0 \\ 
& \textsc{GIN-0} \cite{xu2018powerful} &  \textbf{89.4$\pm$5.6} &  \textbf{64.6$\pm$7.0} & \textbf{76.2$\pm$2.8} &  \textbf{82.7$\pm$1.7} & \textbf{80.2$\pm$1.9} &  \textbf{92.4$\pm$2.5} &  \textbf{57.5$\pm$1.5} \\
& \textsc{GIN-}$\epsilon$ \cite{xu2018powerful} & 89.0$\pm$6.0 & 63.7$\pm$8.2 &  75.9$\pm$3.8 & \textbf{82.7$\pm$1.6} &  80.1$\pm$1.9 &  92.2$\pm$2.3  & 57.0$\pm$1.7 \\
\cmidrule{2-9}

\multirow{5}{*}{\rotatebox{90}{\hspace*{-5pt}   Kernel}}        
& \textsc{GL} \cite{shervashidze2009efficient}  & 81.7$\pm$2.1  & 57.3$\pm$1.4  & -  &  53.9$\pm$0.4 & \textbf{56.3$\pm$0.6}  & 77.3$\pm$0.2 & 41.0$\pm$0.2    \\ 
& \textsc{WL} \cite{shervashidze2011weisfeiler}  & 80.7$\pm$3.0  & 58.0$\pm$0.5 & 72.9$\pm$0.6 &  80.0$\pm$0.5  & -  & 68.8$\pm$0.4 & 46.1$\pm$0.2 \\ 
& \textsc{DGK} \cite{yanardag2015deep} & 87.4$\pm$2.7  & 60.1$\pm$2.6 & \textbf{73.3$\pm$0.8} & \textbf{80.3$\pm$0.5}  & - & \textbf{78.0$\pm$0.4} & 41.3$\pm$0.2  \\ 
& \textsc{MLG} \cite{kondor2016multiscale} & 87.9$\pm$1.6  & \textbf{63.3$\pm$1.5} & 41.2$\pm$0.0 & \textgreater 1 Day  &  \textgreater 1 Day  & 63.3$\pm$1.5 & \textbf{57.3$\pm$1.4} \\ 
& \textsc{GCKN} \cite{chen2020convolutional} & \textbf{87.2$\pm$6.8} & - & 50.8$\pm$0.8 &  70.6$\pm$2.0 &  54.3$\pm$1.0  & 58.4$\pm$7.6 & \textbf{57.3$\pm$1.4} \\ 
\cmidrule{2-9}

\multirow{5}{*}{\rotatebox{90}{\vspace*{13cm} Unsupervised \textbf{ }}} 
& \textsc{Graph2Vec} \cite{narayanan2017graph2vec} & 83.2$\pm$9.3 & 60.2$\pm$6.9 & 73.3$\pm$2.1 & 73.2$\pm$1.8 & - & 75.8$\pm$1.0 & 47.9$\pm$0.3 \\ 
& \textsc{InfoGraph} \cite{sun2019infograph} & 89.0$\pm$1.1 & 61.7$\pm$1.7 & 74.4$\pm$0.3 &  73.8$\pm$0.7 & 67.6$\pm$1.2  & 82.5 $\pm$1.4  & 53.5$\pm$1.0 \\ 
& \textsc{MVGRL} \cite{hassani2020contrastive} & 89.7$\pm$1.1 & 62.5$\pm$1.7 &  -  &  75.0$\pm$0.7 &  68.9$\pm$1.9 & 84.5$\pm$0.6  & - \\   
& \textsc{GCC} \cite{qiu2020gcc} & 86.4$\pm$0.5 & 58.4$\pm$1.2  &  72.9$\pm$0.5 & 66.9$\pm$0.2 &  75.2$\pm$0.3 & 88.4$\pm$0.3 & 52.6$\pm$0.2  \\ 
& \textsc{GraphCL} \cite{you2020graph} & 86.8$\pm$1.3 & 58.4$\pm$1.7  &  74.4$\pm$0.5 & 77.9$\pm$0.4 & 71.4$\pm$1.2 & 89.5$\pm$0.8 & 56.0$\pm$0.3  \\ 
& \textsc{PGCL} (ours)  & \textbf{91.1$\pm$1.2}  & \textbf{63.3$\pm$1.3}  &  \textbf{75.7$\pm$0.2} & \textbf{78.8$\pm$0.8} & \textbf{76.0$\pm$0.3} & \textbf{91.5$\pm$0.7} & \textbf{56.3$\pm$0.2} \\ 

\cmidrule[\heavyrulewidth]{2-9}
\end{tabular}
 \caption{{\bf Graph classification accuracies (\%) of kernel, supervised and unsupervised methods}.  
 We report the mean 10-fold cross-validation accuracy with five runs. ‘\textgreater 1 Day’ represents that the computation exceeds 24 hours.}
  \label{tab:unsup_results}
  \vspace{-0.1in}
\end{table*}

\paragraph{Baselines}
In the unsupervised graph classification, PGCL is evaluated following \cite{sun2019infograph,you2020graph}. We compare our results with five graph kernel
methods including Graphlet Kernel (GL) \cite{shervashidze2009efficient}, Weisfeiler-Lehman Sub-tree Kernel (WL) \cite{shervashidze2011weisfeiler}, Deep Graph Kernels (DGK) \cite{yanardag2015deep}, Multi-Scale Laplacian Kernel (MLG) \cite{kondor2016multiscale} and Graph Convolutional Kernel Network (GCKN\footnote{We report our reproduced results of GCKN for fair comparisons since the original GCKN paper adopts different train-test splits for nested 10-fold cross-validation, which is different from the Stratified10fold splits of other contrastive learning works \cite{sun2019infograph,hassani2020contrastive} and ours.}) \cite{chen2020convolutional}. 
We also compare with four supervised GNNs reported in \cite{xu2018powerful} including GraphSAGE\cite{hamilton2017inductive}, GCN \cite{kipf2016semi} and two variants of GIN \cite{xu2018powerful}: GIN-0 and
GIN-$\epsilon$. Finally, we compare with five unsupervised methods including Graph2Vec \cite{narayanan2017graph2vec}, InfoGraph \cite{sun2019infograph}, MVGRL \cite{hassani2020contrastive}, GCC \cite{qiu2020gcc} and GraphCL \cite{you2020graph}.  We report the results of unsupervised methods based on the released code.

\paragraph{Model Configuration}
We use the \textbf{g}raph \textbf{i}somorphism \textbf{n}etwork (GIN) \cite{xu2018powerful} as the encoder following \cite{you2020graph} to attain node representations for unsupervised graph classification. All projection heads are implemented as two-layer MLPs. For unsupervised graph classification, we adopt LIB-SVM \cite{chang2011libsvm} with \textit{C}  parameter selected in \{$10^{-3}$, $10^{-2}$, \dots, $10^{2}$, $10^{3}$\} as our downstream classifier. Then we use 10-fold cross validation accuracy as the classification performance and repeat the experiments five times to report the mean and standard deviation. We adopt ``node dropping'' and ``edge perturbation'' as the two types of graph augmentations, which perform better than other augmentations (e.g., ``subgraph'') empirically, referring to the implementation\footnote{https://github.com/Shen-Lab/GraphCL}. Prototype vectors are initialized with the default He initialization~\cite{he2015delving} in Pytorch. To help the very beginning of the optimization, we freeze the prototypes during the first few epochs of training and focus on learning the graph representation first. Then the prototype vectors are involved in the optimization of PGCL progressively.
The best hyperparameter $\lambda$ to balance the consistency regularizer and the reweighted contrastive objective is 6. And the number of prototypes is set to 10. The source code of PGCL will be released for reproducibility.

\begin{table*}
\renewcommand\arraystretch{1.03}
\tabcolsep 0.08in
\scriptsize
\centering
\newcommand{\STAB}[1]{\begin{tabular}{@{}c@{}}#1\end{tabular}}
\resizebox{\textwidth}{!}{
\begin{tabular}{lcccccccc}
\cmidrule[\heavyrulewidth]{1-9}
Downstream Dataset    & BBBP & Tox21 & SIDER & ToxCast & ClinTox & BACE & MUV & \multicolumn{1}{c}{\begin{tabular}[c]{@{}c@{}} Average\end{tabular}}  \\
\#Molecules & 2039 & 7831 & 1427 & 8575 & 1478 & 1513  & 93087  & \multicolumn{1}{c}{\begin{tabular}[c]{@{}c@{}} Rank\end{tabular}}\\
\#Tasks & 1 & 12 & 27  & 617 & 2 & 1 & 17 & \multicolumn{1}{c}{\begin{tabular}[c]{@{}c@{}} $(\downarrow)$\end{tabular}}  
\\ \hline
No Pre-Train & 65.8 $\pm$ 4.5 & 74.0 $\pm$ 0.8 & 57.3 $\pm$ 1.6 &  63.4 $\pm$ 0.6 & 58.0 $\pm$ 4.4 & 70.1 $\pm$ 5.4 & 71.8 $\pm$ 2.5 &  6.1  \\ \hline
EdgePred~\cite{hu2020pretraining} & 67.3 $\pm$ 2.4 & 76.0 $\pm$ 0.6 & 60.4 $\pm$ 0.7 & 64.1 $\pm $ 0.6 & 64.1 $\pm$ 3.7 & \textbf{79.9 $\pm$ 0.9} & 74.1 $\pm$ 2.1 & 3.7    \\
AttrMasking~\cite{hu2020pretraining}  & 64.3 $\pm$ 2.8 & \textbf{76.7 $\pm$ 0.4} & 61.0 $\pm$ 0.7 & 64.2 $\pm$ 0.5 & 71.8 $\pm$ 4.1 & 79.3 $\pm$ 1.6 & 74.7 $\pm$ 1.4  & 2.9    \\
ContextPred~\cite{hu2020pretraining}  & 68.0 $\pm$ 2.0 & 75.7 $\pm$ 0.7 & 60.9 $\pm$ 0.6 & 63.9 $\pm$ 0.6 & 65.9 $\pm$ 3.8 & 79.6 $\pm$ 1.2 & \textbf{75.8 $\pm$ 1.7}   & 3.1 \\
InfoGraph~\cite{sun2019infograph}    & 68.8 $\pm$ 0.8 & 75.3 $\pm$ 0.5 & 58.4 $\pm$ 0.8 & 62.7 $\pm$ 0.4 & 69.9 $\pm$ 3.0 & 75.9 $\pm$ 1.6 & 75.3 $\pm$ 2.5    & 4.4 \\
GraphCL~\cite{you2020graph}      & 69.7 $\pm$ 0.7 & 73.9 $\pm$ 0.7 & 60.5 $\pm$ 0.9  & 62.4 $\pm$ 0.6 & \textbf{75.9 $\pm$ 2.7}  & 75.4 $\pm$ 1.4 & 69.8 $\pm$ 2.7   & 5.0 
\\ \hline\hline
PGCL (Ours)   & \textbf{69.8 $\pm$ 1.3} & 75.6 $\pm$ 0.5 & \textbf{61.6 $\pm$ 1.1} & \textbf{66.4 $\pm$ 0.2} & 69.4 $\pm$ 1.4 & 79.3 $\pm$ 1.5 & 71.2 $\pm$ 1.3   & \textbf{2.9} \\
\cmidrule[\heavyrulewidth]{1-9}
\end{tabular}%
}

\caption{\small{Transfer learning performance for chemical molecules property prediction (mean ROC-AUC $\pm$ std. over  10 runs). The best results are highlighted in \textbf{Bold}.} }
\label{tab:transfer_learning}
\end{table*}

\paragraph{Experimental Results}
The results of unsupervised graph level representations for downstream graph classification tasks are presented in Table \ref{tab:unsup_results}. Overall, from the table, we can see that our approach achieves state-of-the-art results with respect to other unsupervised models across all seven datasets. PGCL consistently performs better than unsupervised baselines by considerable margins. For example, on the RDT-B dataset \cite{yanardag2015deep}, it achieves 91.5\% accuracy, i.e., a 2.0\% absolute improvement over previous the state-of-the-art method (GraphCL \cite{you2020graph}). Our model also outperforms graph kernel methods in four out of seven datasets and outperforms the best supervised model in one of the datasets. For example, it harvests a 2.4\% absolute improvement over the state-of-the-art graph kernel method (DGK \cite{yanardag2015deep}) on the PROTEINS dataset. When compared to supervised baselines individually, our model outperforms GraphSAGE in two out of four datasets, and outperforms GCN in three out of seven datasets, e.g., a 5.5\% absolute improvement over GCN on the MUTAG dataset. It is noteworthy that PGCL tightens the gap with respect to the supervised baseline of GIN \cite{xu2018powerful} 
such that their performance gap on four out of seven datasets is less than 2\%. The strong performance verifies the superiority of the proposed PGCL framework.

\subsection{Transfer Learning}
\paragraph{Experimental Setup}
Next, we evaluate the GNN encoders trained by PGCL on transfer learning to predict chemical molecule properties. We follow the setting in [17] and use the same datasets: GNNs are pre-trained on the ZINC-2M dataset using self-supervised learning and later fine-tuned on another downstream dataset to test out-of-distribution performance. We adopt baselines including no pre-trained GIN (i.e., without self-supervised training on the first dataset and with only fine-tuning), InfoGraph \cite{sun2019infograph}, GraphCL \cite{you2021graph} and three different pre-train strategies in \cite{hu2020pretraining} including edge prediction, node attribute masking and context prediction.
Note that \cite{hu2020pretraining} incorporates the domain knowledge heuristically that correlates with the specific downstream datasets.

\paragraph{Experimental Results}
According to Table \ref{tab:transfer_learning}, PGCL significantly outperform all baselines in 3 out of 7 datasets and achieves a mean rank of 2.9 across these 7 datasets. Although without domain knowledge incorporated, PGCL still achieves competitive performance to heuristic self-supervised approaches \cite{hu2020pretraining}. Meanwhile, PGCL outperforms GraphCL \cite{you2021graph} on unseen datasets with better generalizability. In contrast to InfoGraph\cite{sun2019infograph} and GraphCL \cite{you2021graph},  PGCL achieves some performance closer to those heuristic graph pre-training baselines (EdgePred, AttrMasking and ContextPred) based on domain knowledge in \cite{hu2020pretraining}. This is rather significant as our method utilizes only node dropping and edge perturbation as the data augmentation, which again shows the effectiveness of the PGCL.

\begin{table}[h]
\tabcolsep 0.045in
\small
\renewcommand\arraystretch{1.5}
\centering
\scalebox{0.9}
{
\begin{tabular}{cccc|ccccc}
\cmidrule[\heavyrulewidth]{1-8}
$\mathcal{L}_\mathbf{Inf.}$ & $\mathcal{L}_\mathbf{\tiny Con.}$ & $\mathcal{L}_\mathbf{\tiny S.R.}$ & $\mathcal{L}_\mathbf{\tiny P.R.}$   & {\textsc{MUTAG}} & {\textsc{PTC}} & {\textsc{PRO.}} &  {\textsc{COLLAB}} 
 \\  \hline\hline
\textbf{\checkmark}  &  &  &  & 86.8$\pm$1.3  & 58.4$\pm$1.7 & 74.4$\pm$0.5 & 71.4$\pm$1.2\\          
 & \textbf{\checkmark}  &   &  & 89.7$\pm$1.0  & 61.1$\pm$1.7 & 75.4$\pm$0.4 & 71.5$\pm$1.4 \\
 & & \textbf{\checkmark} & &  89.9$\pm$1.1  & 61.9$\pm$0.9 & 73.4$\pm$0.6 & 72.6$\pm$0.5 \\                     
 & & & \textbf{\checkmark} & 90.1$\pm$0.9  & 62.5$\pm$0.7 & 75.2$\pm$0.4 & 73.3$\pm$0.7  \\ \hline 
\textbf{\checkmark} & \textbf{\checkmark} & & & 89.9$\pm$1.0  & 62.4$\pm$2.1 & 75.4$\pm$0.3 & 73.3$\pm$1.2 \\
 & \textbf{\checkmark} & \textbf{\checkmark} & & 91.0$\pm$1.4 & \textbf{63.4$\pm$1.5} & 73.6$\pm$1.1 & 74.6$\pm$0.6  \\ 
 & \textbf{\checkmark} & & \textbf{\checkmark}  & \textbf{91.1$\pm$1.2} & 63.3$\pm$1.3 &  \textbf{75.7$\pm$0.2} & \textbf{76.0$\pm$0.3}  \\ 
\cmidrule[\heavyrulewidth]{1-8}
\end{tabular}
}
\caption{Ablation study for different objective functions on downstream graph classification datasets. As two variants of the vanilla InfoNCE loss, $\mathcal{L}_\mathbf{S.R.}$ denotes caculating the weight in Eqn.~\eqref{eqn:total_self_loss} with the sample distance while $\mathcal{L}_\mathbf{P.R.}$ correspondes to the prototype distance.}
\vspace{-5mm}
\label{tab:ablation}
\end{table}

\subsection{Ablation Studies}
In Table \ref{tab:ablation}, we analyze the effect of various objective functions, including the vanilla InfoNCE loss $\mathcal{L}_\mathbf{Inf.}$, its two variants, i.e., reweighting with prototype distance $\mathcal{L}_\mathbf{P.R.}$ and sample distance $\mathcal{L}_\mathbf{S.R.}$, and the clustering consistency objective $\mathcal{L}_\mathbf{Con.}$. 
When the clustering consistency and the reweighted contrastive objective are individually applied, they perform better than the InfoNCE loss, which benefits from their explorations of the semantic structure of the data. The prototype-based reweighting objective $\mathcal{L}_\mathbf{P.R.}$ outperforms the sample-based one $\mathcal{L}_\mathbf{S.R.}$ in most datasets, since the prototype plays an important role as the pseudo label during negative sampling and provides a more robust reweighting strategy. By simultaneously applying both objectives $\mathcal{L}_\mathbf{Con.}$ and $\mathcal{L}_\mathbf{P.R.}$, our full model (last row) achieves better performance than merely combining the InfoNCE loss and the clustering consistency, which indicates that the prototype-based reweighting strategy can mitigate the sampling bias problem. 

\begin{figure}
\centering
\begin{center}
\label{fig:num_prototypes}
\centering
\includegraphics[width=6cm]{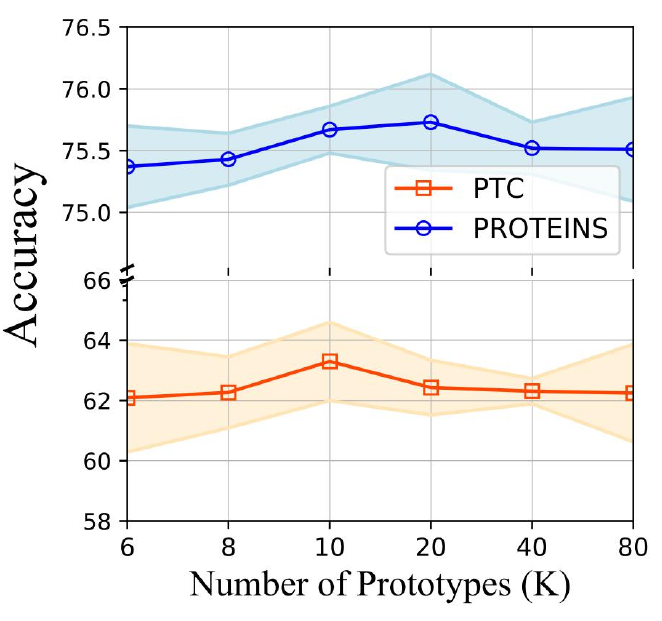}
\caption{\textbf{Sensitivity analysis for the number of prototypes $K$.}}
\label{fig:nmb_proto}
\vspace{-5mm}
\end{center}
\end{figure}

\begin{figure}
\centering
\begin{center}
\label{fig:batch_size}
\centering
\includegraphics[width=6cm]{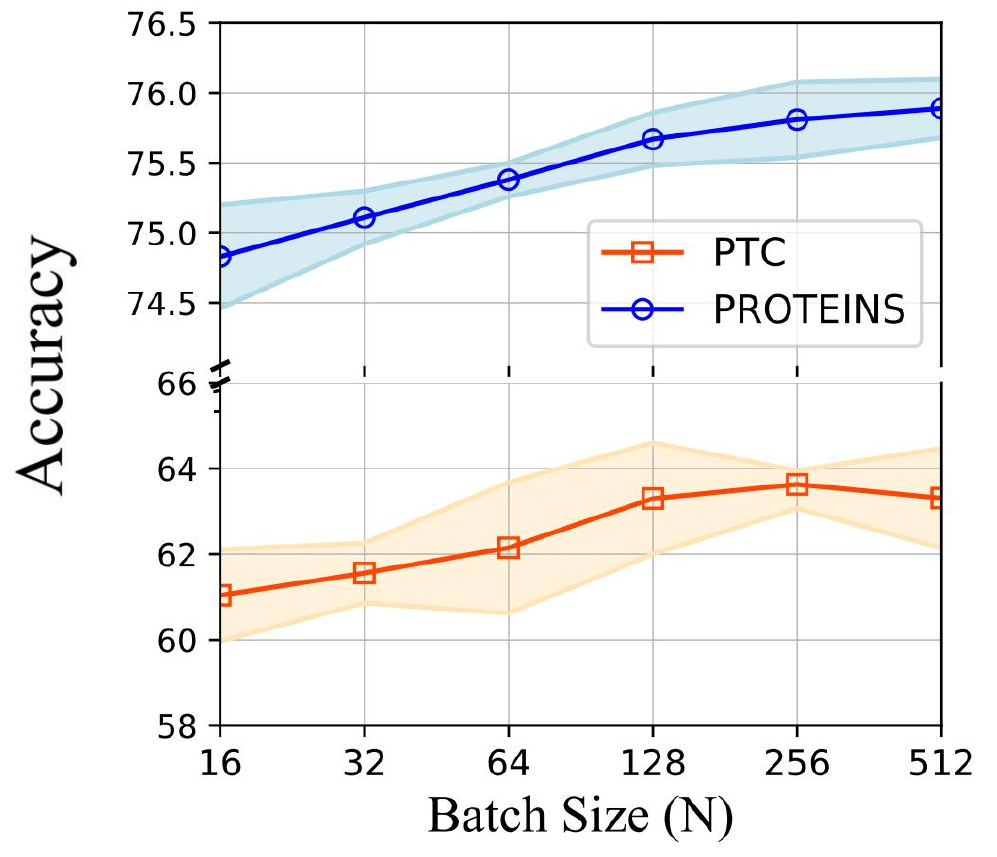}
\caption{\textbf{Sensitivity analysis for batch size $N$.}}
\label{fig:batch_size}
\vspace{-5mm}
\end{center}
\end{figure}

\subsection{Sensitivity Analysis}
\paragraph{Sensitivity to prototype numbers $K$}
In this part, we discuss the selection of parameter $K$ which is the number of prototypes (clusters). Fig. \ref{fig:nmb_proto} shows the performance of PGCL with different number of prototypes $K$ from 6 to 80 on PTC and PROTEINS. It can be observed that at beginning, increasing the number of prototypes improves the performance while too many prototypes leads to slight performance drop, which we conjecture that the model degenerates to the case without prototypes (i.e., each graph acts as a prototype itself). Overall, our PGCL is robust to the prototype numbers.   


\begin{figure*}[t]
\scalebox{1.0}{
    \centering
    \includegraphics[width=\linewidth]{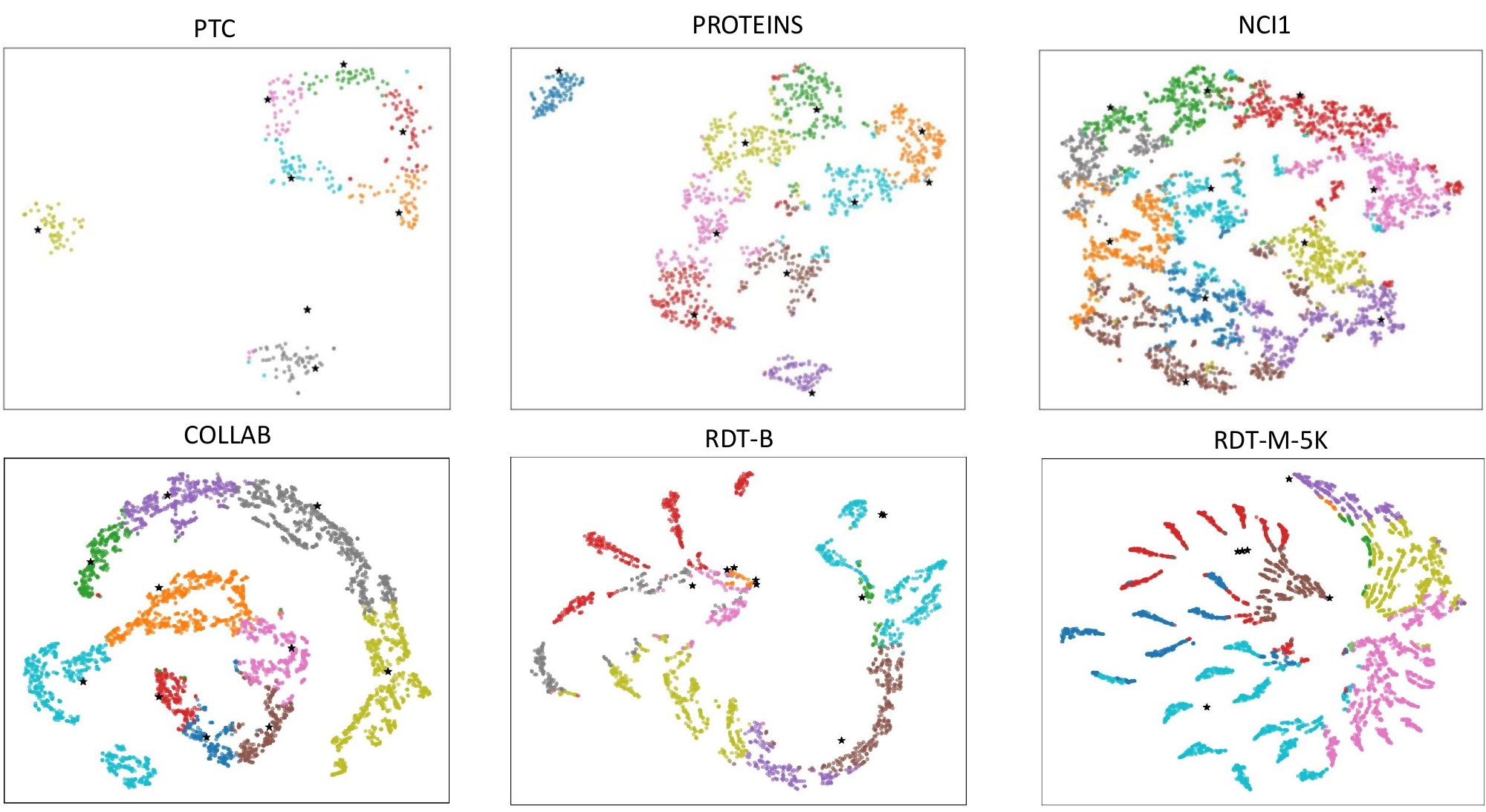}} %
    \caption{\textbf{T-SNE visualization of the learned representation on six datasets.} ``$\star$'' means the prototype vectors. Colors represent underlying classes that PGCL discovers.}
    \label{fig:visualization}
    \vspace{-10pt}
\end{figure*}

\paragraph{Sensitivity to batch size $N$}
In this experiment, we evaluate the effect of batch size $N$ on our model performance. Fig. \ref{fig:batch_size} shows the classification accuracy of our model using different batch sizes from 32 to 512 on PTC and PROTEINS. From the line chart, we can observe that a large batch size (i.e. $N$ > 32) can consistently improve the performance of PGCL. This observation aligns with the case in the image domain \cite{chen2020simple}. 


\subsection{Visualization Results.}
In Fig. \ref{fig:visualization}, we utilize the t-SNE \cite{van2008visualizing} to visualize the graph representations and prototype vectors with the number of clusters $K = 10$ on various datasets. Generally, the representations learned by the proposed PGCL forms separated clusters, where the prototypes fall into the center. It demonstrates that PGCL can discover the underlying global semantic structure over the entire data distribution. Moreover, it can be observed that each cluster has a similar number of samples, which indicates the effectiveness of the equal-partition constraints during clustering.


\section{Conclusions}
We proposed a clustering-based approach called PGCL for unsupervised graph-level representation learning. PGCL clusters semantically similar graphs into the same group, and simultaneously  encourages the clustering consistency for different graph views. Moreover, to mitigate the sampling bias issue, PGCL reweights its negative pairs based on the distance between their prototypes. Benefiting from modeling the global semantic structure via clustering, we achieve new state-of-the-art performance compared to previous unsupervised learning methods on seven graph classification benchmarks.

\bibliographystyle{unsrt}




\begin{IEEEbiography}[{\includegraphics[width=1in,height=1.1in,clip,keepaspectratio]{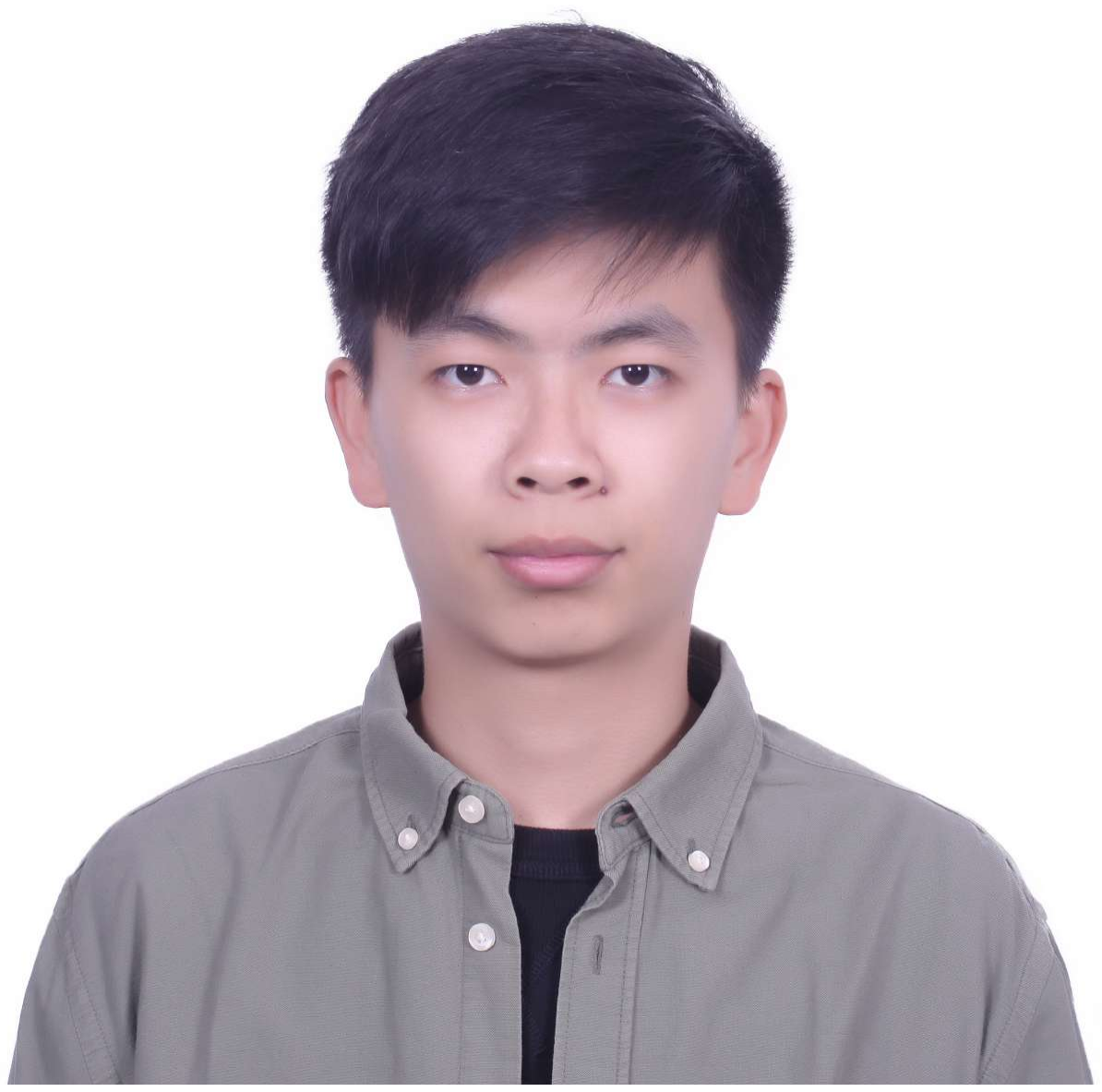}}]{Shuai Lin} received the bachelor’s degree in communication engineering from Xidian University, Xi'an, China, in 2019.  He is currently pursuing the master's degree under the supervision of Xiaodan Liang, with the Department of Intelligent Engineering, Sun Yat-sen University, Guangzhou, China. His main research interests include data mining and interpretable machine learning.
\end{IEEEbiography}

\begin{IEEEbiography}[{\includegraphics[width=1in,height=1.1in,clip,keepaspectratio]{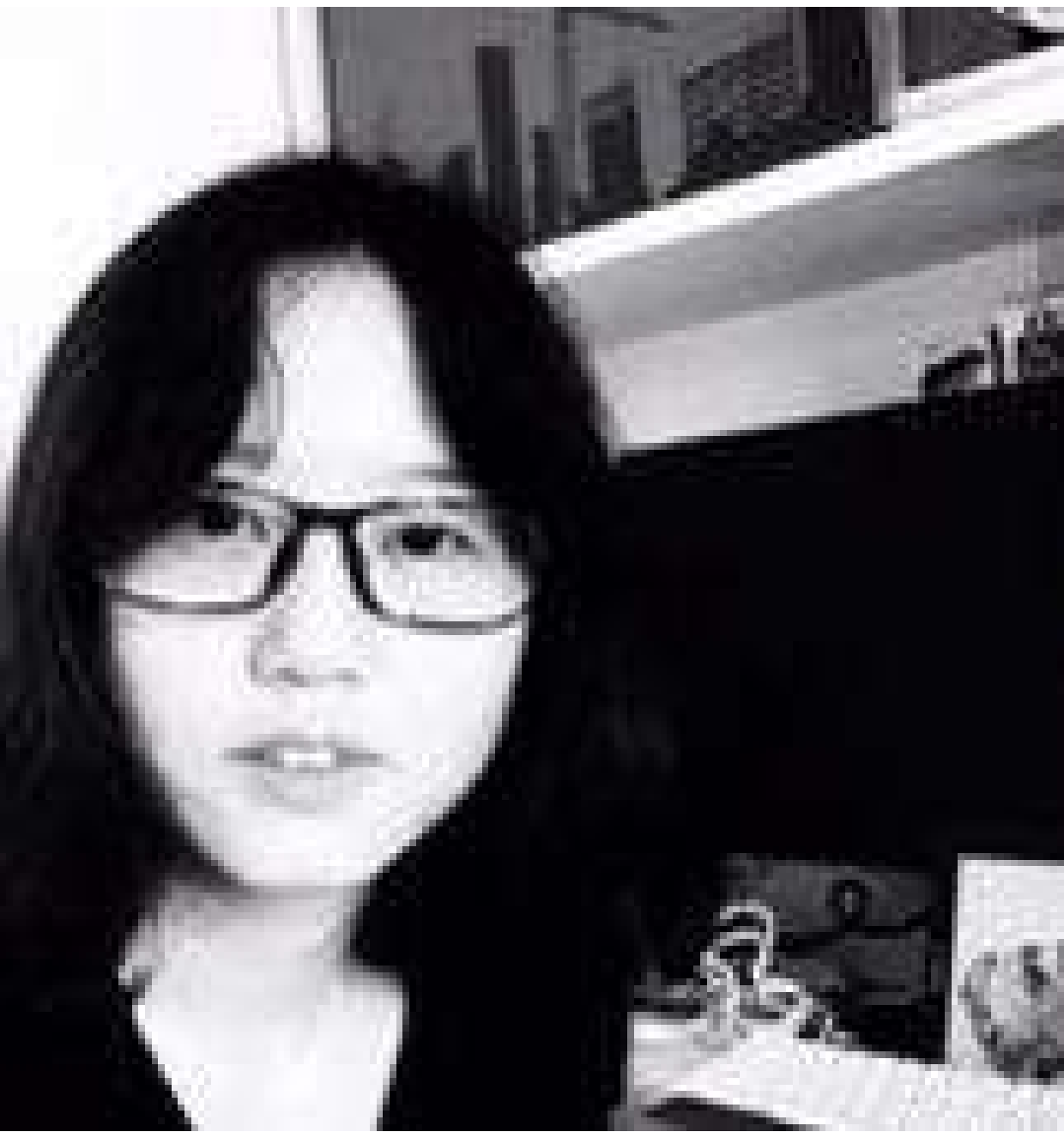}}]{Chen Liu} received the bachelor’s (2013) and master's (2016) degree in psycholoy from Nanjing University, Nanjing, China and University of York, York, United Kingdom respectively.  She is currently pursuing the Ph.d.'s degree under the supervision of Xiaodan Liang, with the Department of Intelligent Engineering, Sun Yat-sen University, Guangzhou, China. Her research interests include deep learning, graph-structured data mining, and their application in timeneural signals etc.
\end{IEEEbiography}

\begin{IEEEbiography}[{\includegraphics[width=1in,height=1.1in,clip,keepaspectratio]{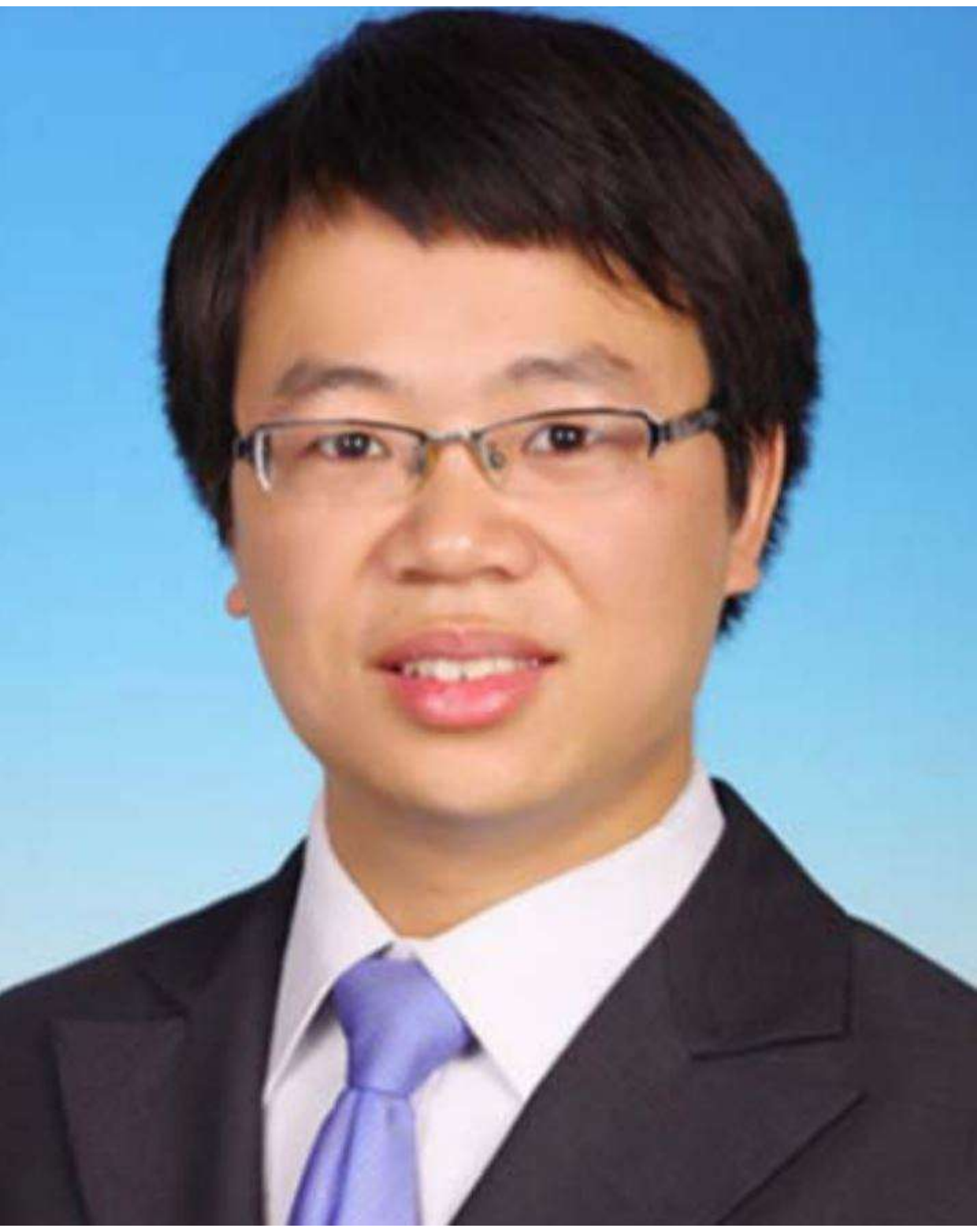}}]{Pan Zhou}
received the master’s degree in computer
science from Peking University, Beijing, China, in 2016, and the Ph.D. degree in computer science from the National University of Singapore, Singapore, in 2019.
He is currently the Senior Research Scientist of the SEA AI Laboratory, SEA Group, Singapore. From 2019 to 2020, he was the Research Scientist of Salesforce, Singapore. His research interests include computer vision, machine learning, and optimization.
Dr. Zhou was the winner of the Microsoft Research Asia Fellowship in 2018.
\end{IEEEbiography}

\begin{IEEEbiography}[{\includegraphics[width=1in,height=1.1in,clip,keepaspectratio]{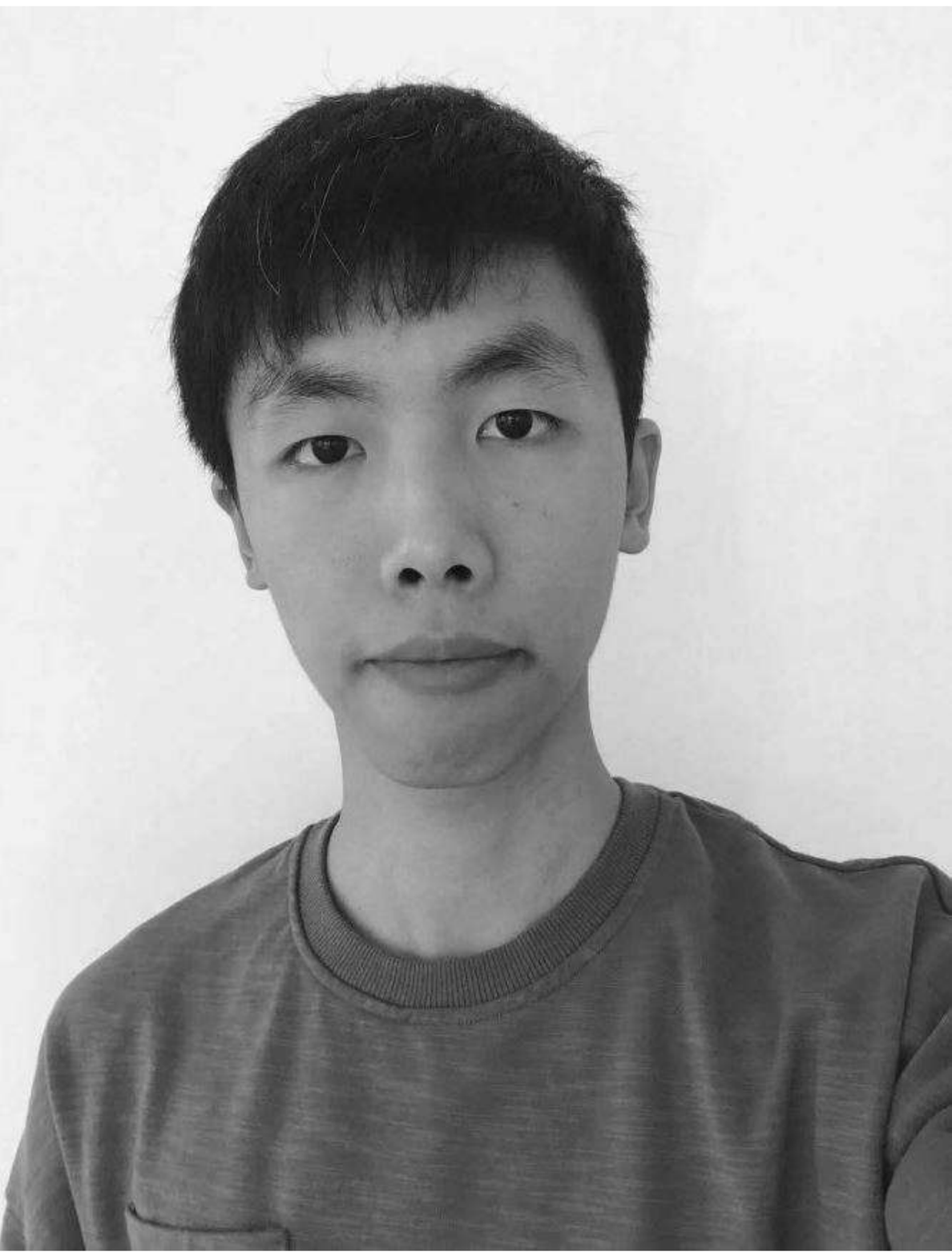}}]{Zi-Yuan Hu} is currently an undergraduate student in computer
science from Sun Yat-sen University, Guangzhou, China. He is good at algorithm design and implementation. His research interest is data mining and cross-modality pre-training.
\end{IEEEbiography}

\begin{IEEEbiography}[{\includegraphics[width=1in,height=1.25in,clip,keepaspectratio]{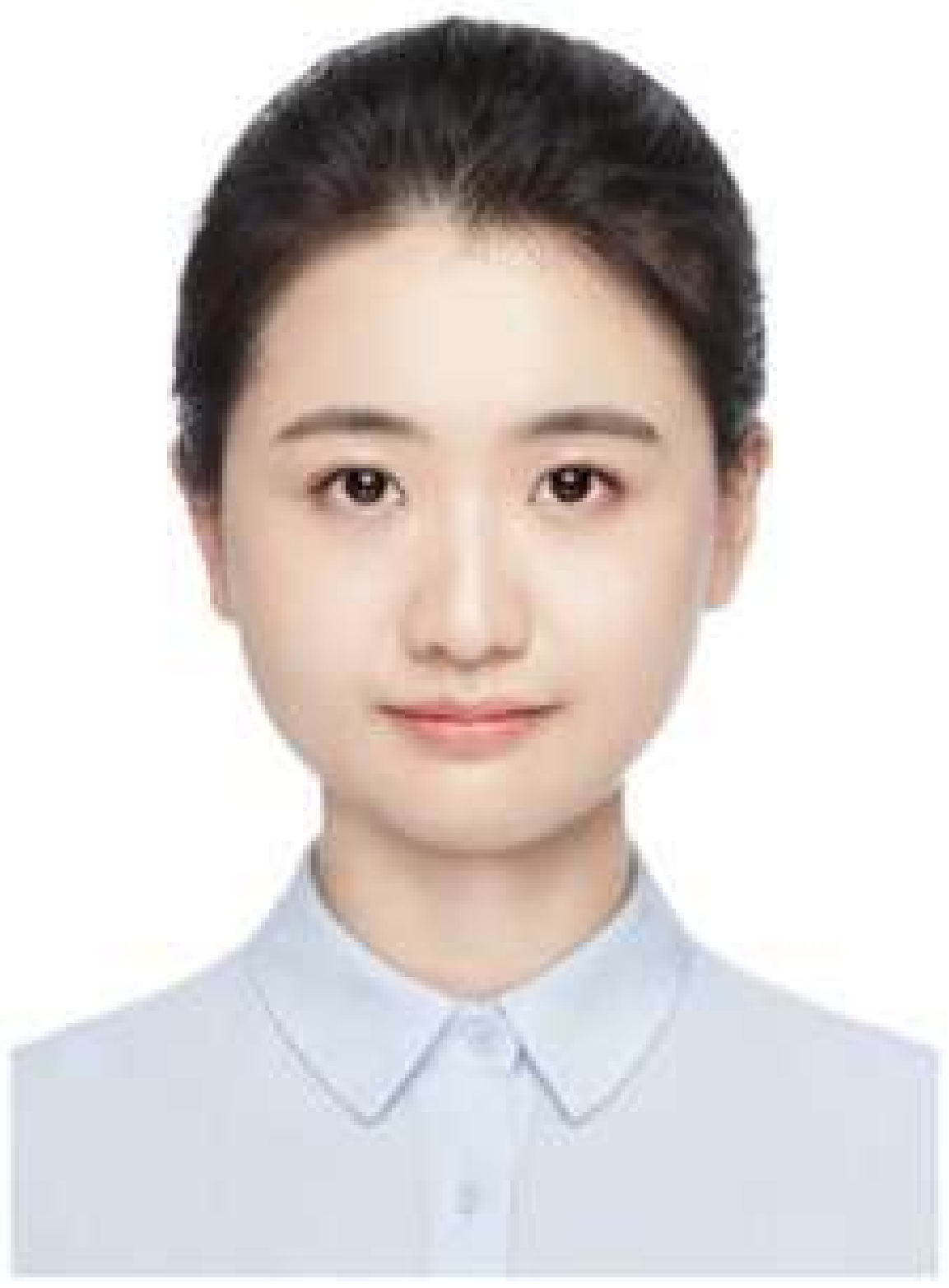}}]{Shuojia Wang} received the Ph.D. degree in Epidemiology and Health Statistics from Zhejiang University, Hangzhou, China in 2020. She is currently the data scientist in Tencent Jarvis lab. Her research interests include data mining, medical decision-making, and disease prediction.
\end{IEEEbiography}

\begin{IEEEbiography}[{\includegraphics[width=1in,height=1.25in,clip,keepaspectratio]{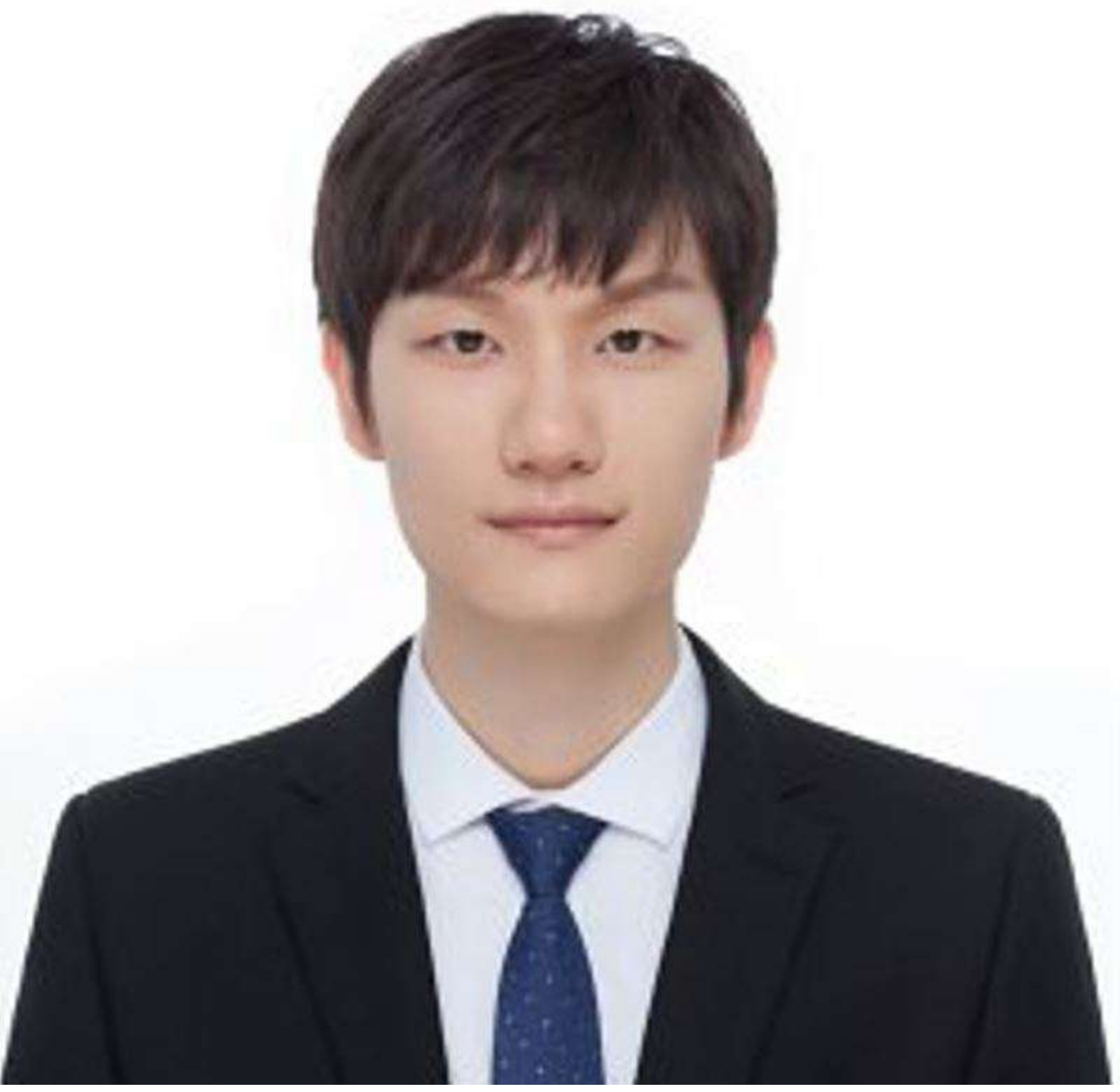}}]{Ruihui Zhao} received his B.S. degree at UESTC (2015) and M.S. degree at Waseda University (2017). He joined Tencent Jarvis Lab as a senior researcher in early 2018. Previously, he was an NLP engineer at Sinovation Ventures (2017 - 2018). He has several papers accepted by ACL, WWW, AAAI, NAACL, IJCNN, TOIT, IEEE WCSP, etc. His research and projects mainly focus on NLP and information security.
\end{IEEEbiography}

\begin{IEEEbiography}[{\includegraphics[width=1in,height=1.25in,clip,keepaspectratio]{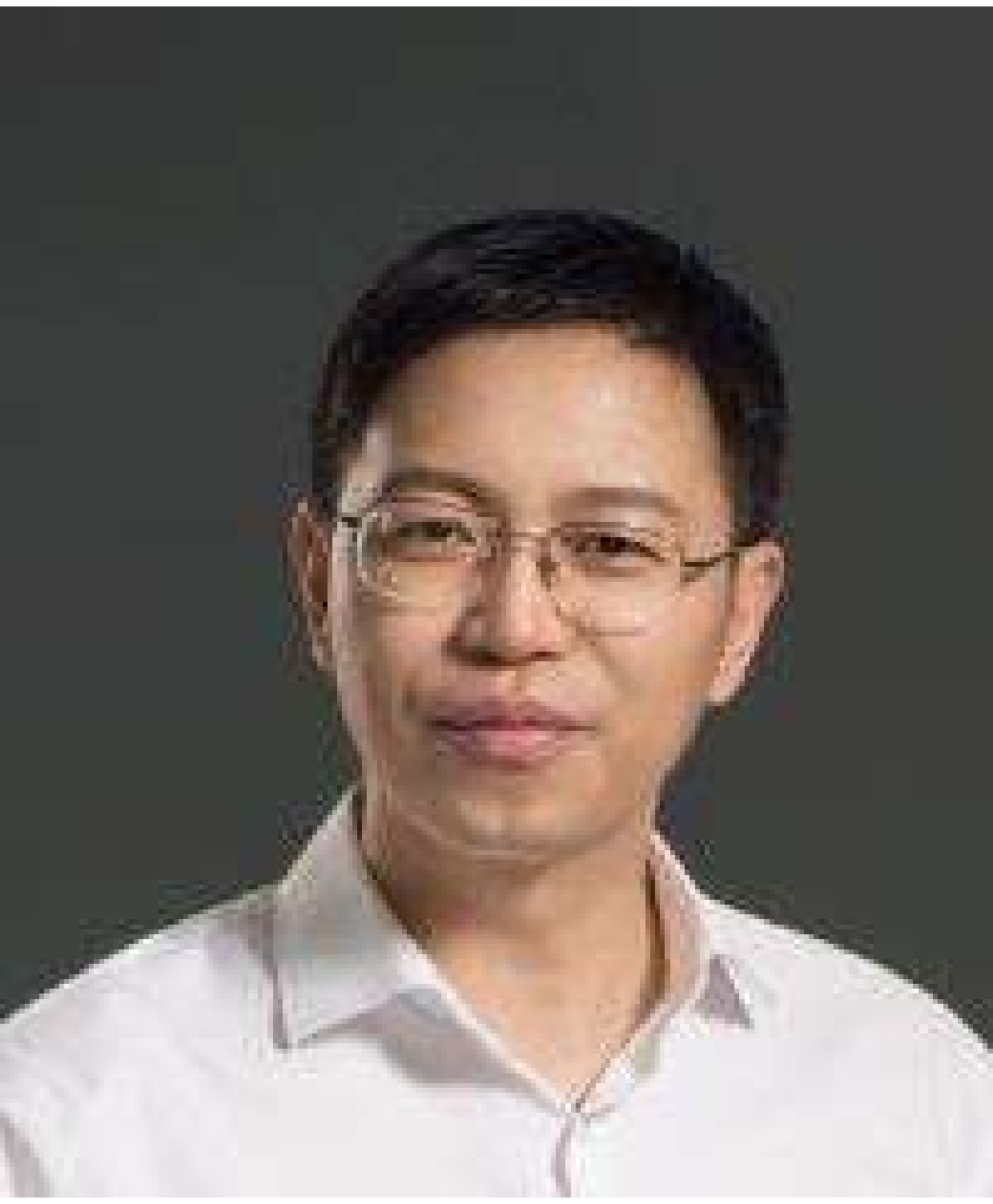}}]{Yefeng Zheng} (Fellow, IEEE) received the BE and ME degrees from Tsinghua University, Beijing, in 1998 and 2001, respectively, and the PhD degree from the University of Maryland, College Park, in 2005. After graduation, he joined Siemens Corporate Research in Princeton, New Jersey, USA. He is now Director and Distinguished Scientist of Tencent Jarvis Lab, Shenzhen, China, leading the company's initiative on Medical AI.
His research interests include medical image analysis, graph data mining and deep learning. 
He is a fellow of the Institute of Electrical and Electronics Engineers (IEEE), a fellow of American Institute for Medical and Biological Engineering (AIMBE).
\end{IEEEbiography}

\begin{IEEEbiography}[{\includegraphics[width=1in,height=1.25in,clip,keepaspectratio]{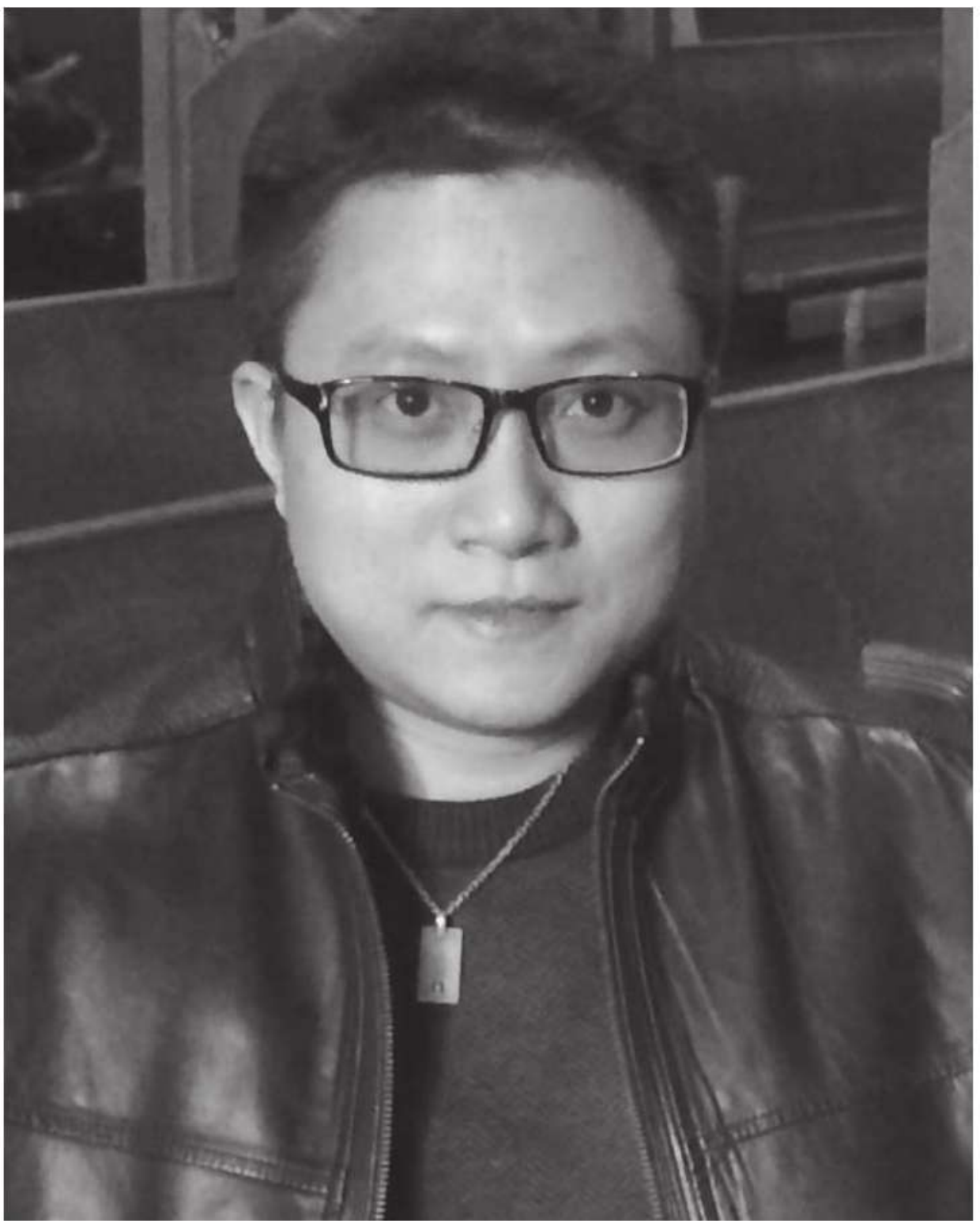}}]{Liang Lin}
(Senior Member, IEEE) served as the
Executive Director of the SenseTime Group, Hong Kong, from 2016 to 2018, leading the research and development teams in developing cutting-edge, deliverable solutions in computer vision and data mining. He is currently a Full Professor of computer science with Sun Yat-sen University, Guangzhou.
He has authored or coauthored more than 200 articles in leading academic journals and conferences, such as IEEE TRANSACTIONS ON PATTERN ANALYS IS AND MACHINE INTELLIGENCE (TPAMI), Conference on Neural Information Processing Systems (NeurIPS), International Conference on Machine Learning (ICML), and The Association
for the Advancement of Artificial Intelligence (AAAI).
Dr. Lin is a fellow of IET. He served as the Area/Session Chair for numerous
conferences, such as CVPR, ICME, ICCV, and International Conference
on Multimedia Retrieval (ICMR). He is an Associate Editor of the IEEE Transactions on Neural Networks and Learning Systems (IEEE TNNLS).
\end{IEEEbiography}

\begin{IEEEbiography}[{\includegraphics[width=1in,height=1.25in,clip,keepaspectratio]{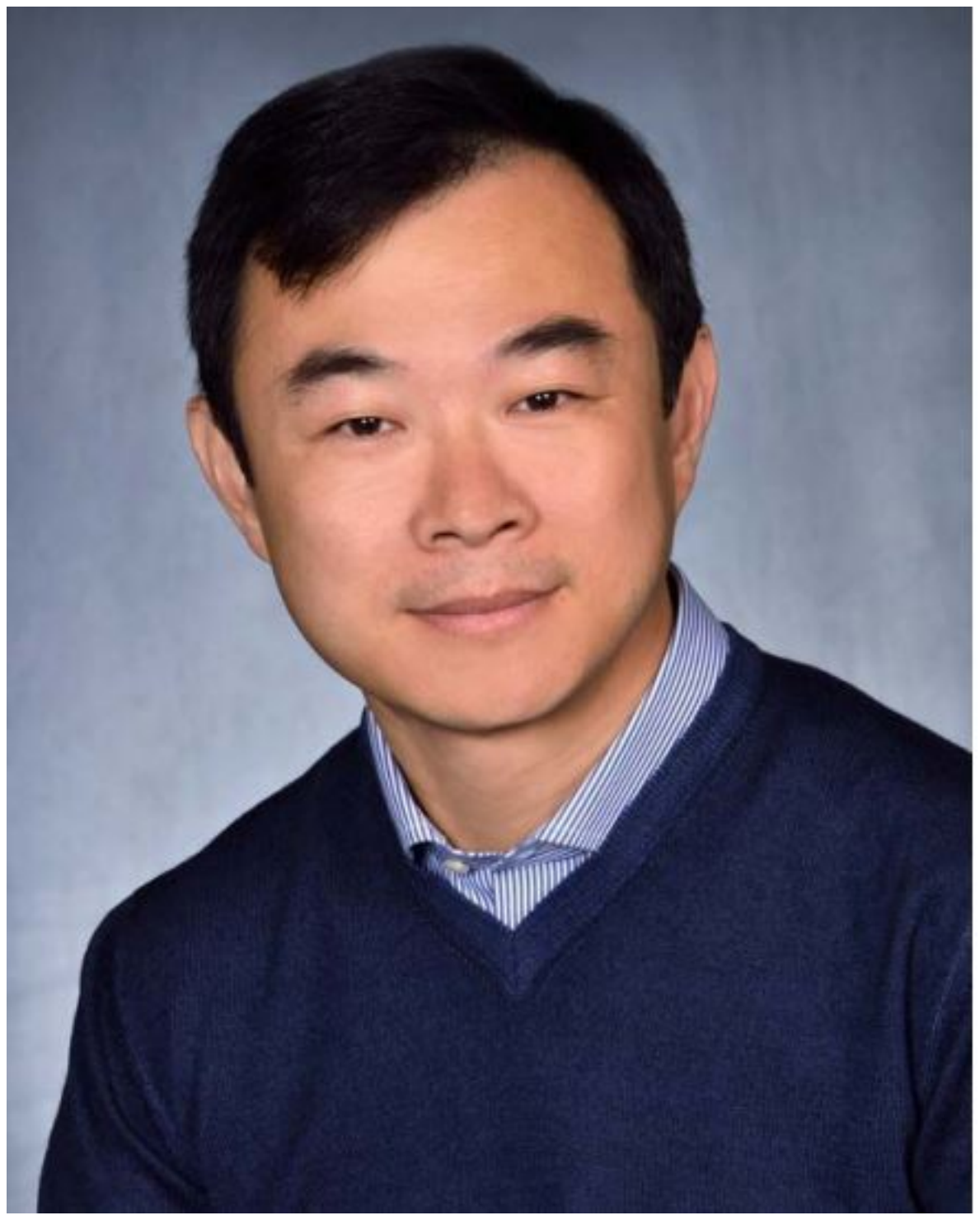}}]{Eric Xing}
(Fellow, IEEE) received the Ph.D. degree
in molecular biology from Rutgers University, New Brunswick, NJ, USA, in 1999, and the Ph.D.
degree in computer science from the University of California at Berkeley, Berkeley, CA, USA, in 2004. He is currently a Professor of machine learning with the School of Computer Science and the Director of the CMU Center for Machine Learning and Health, Carnegie Mellon University, Pittsburgh, PA, USA. His principal research interests lie in the development of machine learning and statistical methodology, especially for solving problems involving automated learning, reasoning, and decision-making in high-dimensional, multimodal, and dynamic possible worlds in social and biological systems.
Dr. Xing is a member of the DARPA Information Science and Technology (ISAT) Advisory Group and the Program Chair of the International Conference on Machine Learning (ICML) 2014. He is also an Associate Editor of the IEEE TRANSACTIONS ON PATTERN ANALYS IS AND MACHINE INTELLIGENCE (PAMI) and the Machine Learning Journal (MLJ) and the Journal of Machine Learning Research (JMLR).
\end{IEEEbiography}

\begin{IEEEbiography}[{\includegraphics[width=1in,height=1.25in,clip,keepaspectratio]{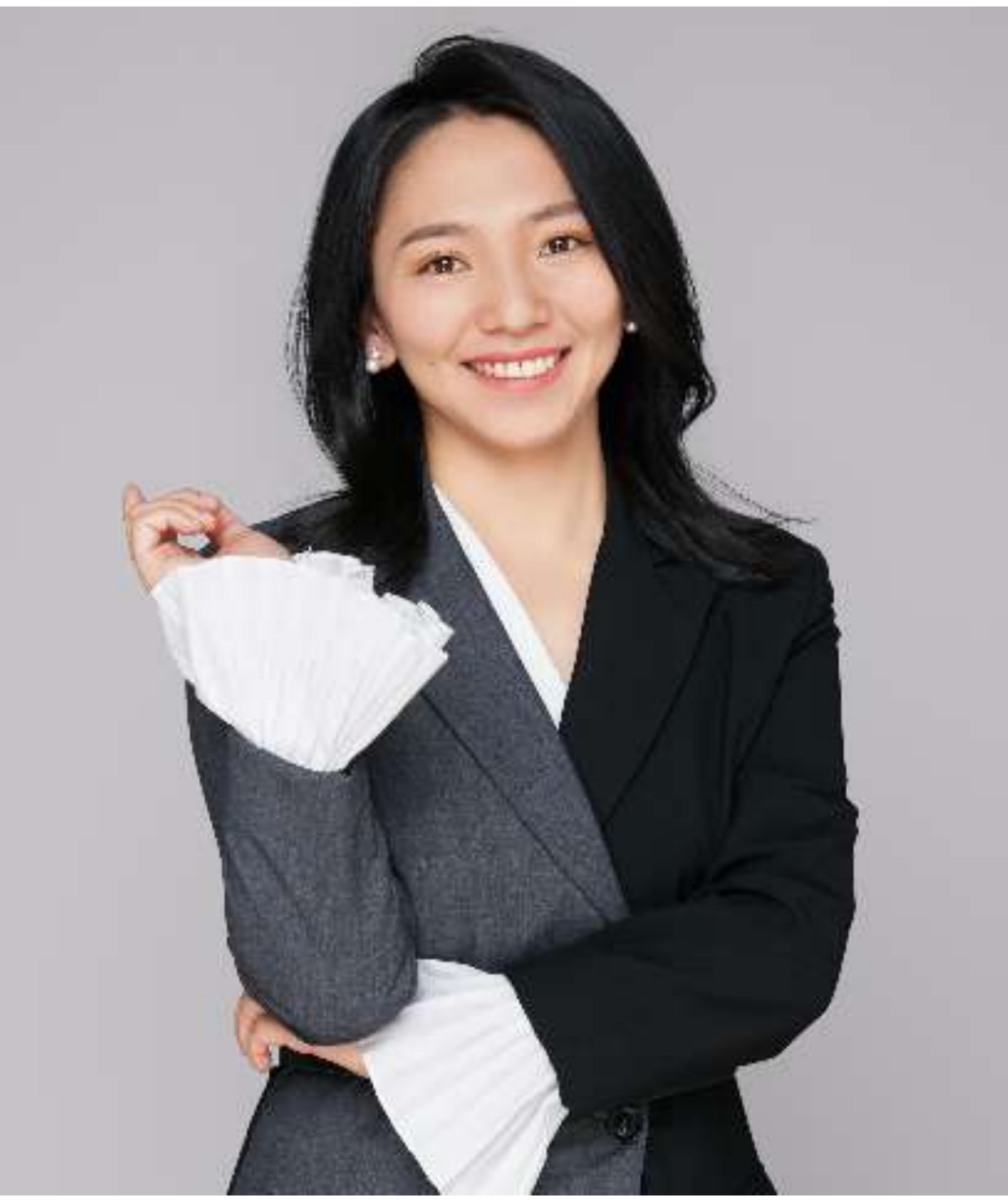}}]{Xiaodan Liang}
(Senior Member, IEEE) received the Ph.D.
degree from Sun Yat-sen University, Guangzhou,
China, in 2016, under the supervision of Liang
Lin. She was a Post-Doctoral Researcher with the Department of Machine Learning, Carnegie Mellon University, Pittsburgh, PA, USA, working with Prof. Eric Xing from 2016 to 2018. She is currently an Associate Professor with Sun Yat-sen University.
She has authored or coauthored several cutting-edge projects on interpretable machine learning, data mining and graph neural network.
\end{IEEEbiography}

\end{document}